\documentclass{article} 
\usepackage{cancel}
\usepackage{iclr2017_conference,times}
\usepackage{hyperref}
\usepackage{url}
\usepackage{algorithm,algpseudocode}
\usepackage{amsfonts}
\usepackage{amsmath}
\usepackage{bm}
\usepackage{graphicx}
\usepackage{subcaption}
\usepackage[page]{appendix}
\graphicspath{ {./images/} }

\title{Adversarial Machine Learning at Scale}

\author{Alexey Kurakin \\
Google Brain \\
\texttt{kurakin@google.com}
\And
Ian J. Goodfellow \\
OpenAI \\
\texttt{ian@openai.com}
\And
Samy Bengio \\
Google Brain \\
\texttt{bengio@google.com}
}

\DeclareMathOperator*{\argmin}{arg\,min}

\DeclareMathOperator{\sign}{sign}

\iclrfinalcopy 

\begin{document}

\maketitle

\begin{abstract}
Adversarial examples are malicious inputs designed to fool machine learning models.
They often transfer from one model to another, allowing attackers to mount black
box attacks without knowledge of the target model's parameters.
Adversarial training is the process of explicitly training a model on adversarial
examples, in order to make it more robust to attack or to reduce its test error
on clean inputs.
So far, adversarial training has primarily been applied to small problems.
In this research, we apply adversarial training to ImageNet~\citep{russakovsky2014imagenet}.
Our contributions include:
(1) recommendations for how to succesfully scale adversarial training to large models and datasets,
(2) the observation that adversarial training confers robustness to single-step attack methods,
(3) the finding that multi-step attack methods are somewhat less transferable than single-step attack
      methods, so single-step attacks are the best for mounting black-box attacks,
      and
(4) resolution of a ``label leaking'' effect that causes adversarially trained models to perform
      better on adversarial examples than on clean examples, because the adversarial
      example construction process uses the true label and the model can learn to
      exploit regularities in the construction process.
\end{abstract}

\section{Introduction}

It has been shown that machine learning models are often vulnerable to 
adversarial manipulation of their input intended to cause
incorrect classification \citep{dalvi2004adversarial}.
In particular, neural networks and many other categories of machine
learning models are highly vulnerable to
attacks based on small modifications of the input to the model
at test time
\citep{biggio2013evasion,Szegedy-ICLR2014,Goodfellow-2015-adversarial,Papernot-2016-transferability}.

The problem can be summarized as follows.
Let's say there is a machine learning system $M$ and input sample $C$ which we call a clean example.
Let's assume that sample $C$ is correctly classified by the machine learning system, i.e. $M(C) = y_{true}$.
It's possible to construct an adversarial example $A$ which is perceptually indistinguishable 
from $C$ but is classified incorrectly, i.e. $M(A) \ne y_{true}$.
These adversarial examples are misclassified far more often than examples that have been perturbed
by noise, even if the magnitude of the noise is much larger than the magnitude of the adversarial
perturbation \citep{Szegedy-ICLR2014}.

Adversarial examples pose potential security threats for practical machine learning applications.
In particular, \citet{Szegedy-ICLR2014} showed that an adversarial example that was designed
to be misclassified by a model $M_1$ is often also misclassified by a model $M_2$.
This adversarial example transferability property means 
that it is possible to generate adversarial examples
and perform a misclassification attack on a machine learning system without access
to the underlying model.
\citet{Papernot-MGJCS16} and \citet{Papernot-2016-transferability} demonstrated such
attacks in realistic scenarios.

It has been shown~\citep{Goodfellow-2015-adversarial,LearningWithStrongAdversary} that
injecting adversarial examples into the training set (also called adversarial training) could
increase robustness of neural networks to adversarial examples.
Another existing approach is to use defensive distillation to train the network~\citep{PapernotDistillation2015}.
However all prior work studies defense measures only on relatively small datasets like MNIST and CIFAR10.
Some concurrent work studies attack mechanisms on ImageNet \citep{rozsa2016accuracy},
focusing on the question of how well adversarial examples transfer between different
types of models, while we focus on defenses and studying how well different types
of adversarial example generation procedures transfer between relatively similar models.

In this paper we studied adversarial training of Inception models trained on ImageNet.
The contributions of this paper are the following:
\begin{itemize}
\item We successfully used adversarial training to train an Inception v3 model~\citep{Inception-v3}
on ImageNet dataset~\citep{russakovsky2014imagenet} and to significantly increase
robustness against adversarial examples generated by the \textit{fast gradient sign method}~\citep{Goodfellow-2015-adversarial}
as well as other one-step methods.
\item We demonstrated that different types of adversarial examples tend to
have different transferability properties between models.
In particular we observed that
those adversarial examples which are harder to resist using adversarial training
are less likely to be transferrable between models.
\item We showed that models which have higher capacity (i.e. number of parameters) tend to
be more robust to adversarial examples compared to lower capacity model of the same architecture.
This provides additional cue which could help building more robust models.
\item We also observed an interesting property we call ``label leaking''.
Adversarial examples constructed with a single-step method making use of the true labels
may be easier to classify than clean adversarial examples, because an adversarially
trained model can learn to exploit regularities in the adversarial example construction
process.
This suggests using adversarial example construction processes that do not make use of
the true label.
\end{itemize}

The rest of the paper is structured as follows:
In section~\ref{sec:adv-examples} we review different methods to generate adversarial examples.
Section~\ref{sec:adv-training} describes details of our adversarial training algorithm.
Finally, section~\ref{sec:experiments} describes our experiments and results of adversarial training.

\section{Methods generating adversarial examples}\label{sec:adv-examples}

%
%
%
%

\subsection{Terminology and Notation}

In this paper we use the following notation and terminology regarding adversarial examples:

\begin{enumerate}
  \item $\bm{X}$, the \textit{clean image}~--- unmodified image from the dataset (either train or test set).
  \item $\bm{X}^{adv}$, the \textit{adversarial image}: the output of any procedure intended to produce an approximate
    worst-case modification of the clean image. We sometimes call this a
    \textit{candidate adversarial image} to emphasize that an adversarial image is not
    necessarily misclassified by the neural network.
  \item \textit{Misclassified adversarial image}~--- candidate adversarial image which is misclassified by the neural network.
  In addition we are typically interested only in those misclassified adversarial images when
  the corresponding clean image is correctly classified.
\item $\epsilon$: The size of the adversarial perturbation. In most cases, we require the
  $L_\infty$ norm of the perturbation to be less than $\epsilon$, as done by
  \citet{Goodfellow-2015-adversarial}.
  We always specify $\epsilon$ in terms of pixel values in the range $[0, 255]$.
  Note that some other work on adversarial examples minimizes the size of the perturbation
  rather than imposing a constraint on the size of the perturbation~\citep{Szegedy-ICLR2014}.
\item The cost function used to train the model is denoted $J(\bm{X}, y_{true})$.
\item $Clip_{\bm{X}, \epsilon}( \bm{A} )$ denotes element-wise clipping $\bm{A}$, with $A_{i,j}$
  clipped to the range $[X_{i,j} - \epsilon, X_{i,j} + \epsilon ]$.
\item {\em One-step} methods of adversarial example generation generate a candidate
      adversarial image after computing only one gradient.
      They are often based on finding the optimal perturbation of a linear approximation
      of the cost or model.
      {\em Iterative} methods apply many gradient updates. They typically do not rely
      on any approximation of the model and typically produce more harmful adversarial
      examples when run for more iterations.
\end{enumerate}

\subsection{Attack methods}

We study a variety of attack methods:

\paragraph{Fast gradient sign method}\label{sec:fast-method}

\citet{Goodfellow-2015-adversarial} proposed the \textit{fast gradient sign method} (FGSM)
as a simple way to generate adversarial examples:
\begin{equation}
\label{eq:fast-adv-examples}
\bm{X}^{adv} = \bm{X} + \epsilon \sign \bigl( \nabla_X J(\bm{X}, y_{true})  \bigr)
\end{equation}
This method is simple and computationally efficient compared to more complex methods like L-BFGS~\citep{Szegedy-ICLR2014},
however it usually has a lower success rate.
On ImageNet, top-1 error rate on candidate adversarial images for the FGSM is about $63\% - 69\%$ for $\epsilon \in [2, 32]$.

\paragraph{One-step target class methods}\label{sec:step-ll-method}

FGSM finds adversarial perturbations which increase the value of the loss function.
An alternative approach is to maximize probability $p(y_{target} \mid \bm{X})$
of some specific target class $y_{target}$ which is unlikely to be the true class for a given image.
For a neural network with cross-entropy loss this will lead to the following formula for the one-step target class method:
\begin{equation}
\label{eq:stepll-adv-examples}
\bm{X}^{adv} = \bm{X} - \epsilon \sign \bigl( \nabla_X J(\bm{X}, y_{target}) \bigr)
\end{equation}

As a target class we can use the least likely class predicted by the network
$y_{LL} = \argmin_{y} \bigl\{ p( y \mid \bm{X} ) \bigr\}$,
as suggested by \citet{PhysicalAdversarialExamples}.
In such case we refer to this method as one-step least likely class or just ``step l.l.''
Alternatively we can use a random class as target class.
In such a case we refer to this method as ``step rnd.''.

\paragraph{Basic iterative method}

A straightforward extension of FGSM is to apply it multiple times with small step size:

\[
\bm{X}^{adv}_{0} = \bm{X}, \quad
\bm{X}^{adv}_{N+1} = Clip_{X, \epsilon}\Bigl\{ \bm{X}^{adv}_{N} + \alpha \sign \bigl( \nabla_X J(\bm{X}^{adv}_{N}, y_{true})  \bigr) \Bigr\}
\]

In our experiments we used $\alpha = 1$, i.e. we changed the value of each pixel only by $1$ on each step.
We selected the number of iterations to be $\min(\epsilon + 4, 1.25\epsilon)$.
See more information on this method in~\citet{PhysicalAdversarialExamples}.
Below we refer to this method as ``iter. basic'' method.

\paragraph{Iterative least-likely class method}

By running multiple iterations of the ``step l.l.'' method we can get adversarial examples which are misclassified in more than $99\%$ of the cases:
\[
\bm{X}^{adv}_{0} = \bm{X}, \quad
\bm{X}^{adv}_{N+1} = Clip_{X, \epsilon}\left\{ \bm{X}^{adv}_{N} - \alpha \sign \left( \nabla_X J(\bm{X}^{adv}_{N}, y_{LL})  \right) \right\}
\]

$\alpha$ and number of iterations were selected in the same way as for the basic iterative method.
Below we refer to this method as the ``iter. l.l.''.

\section{Adversarial training}\label{sec:adv-training}

The basic idea of adversarial training is to inject adversarial examples into the training
set, continually generating new adversarial examples at every step of training \citep{Goodfellow-2015-adversarial}.
Adversarial training was originally developed for small models that did not use
batch normalization.
To scale adversarial training to ImageNet, we recommend using batch
normalization \citep{Ioffe+Szegedy-2015}.
To do so successfully, we found that it was important for
examples to be grouped into batches containing both normal and adversarial examples before
taking each training step, as described in algorithm~\ref{alg:adv-training}.

\begin{algorithm}
\caption{Adversarial training of network $N$.\\
Size of the training minibatch is $m$. Number of adversarial images in the minibatch is $k$.}\label{alg:adv-training}
\begin{algorithmic}[1]
\State Randomly initialize network $N$
\Repeat
  \State Read minibatch $B = \{ X^1, \ldots, X^m \}$ from training set
  \State Generate $k$ adversarial examples $\{ X^1_{adv}, \ldots, X^k_{adv} \}$ from corresponding
  \Statex \hspace{\algorithmicindent} clean examples $\{ X^1, \ldots, X^k \}$ using current state of the network $N$
  \State Make new minibatch $B' = \{ X^1_{adv}, \ldots, X^k_{adv}, X^{k+1}, \ldots, X^m \}$
  \State Do one training step of network $N$ using minibatch $B'$
\Until training converged
\end{algorithmic}
\end{algorithm}

We use a loss function that allows independent control of the number and relative weight
of adversarial examples in each batch:
\[
Loss = \frac{1}{(m-k) + \lambda k}\left( \sum_{i \in CLEAN} L(X_{i}|y_{i}) + \lambda \sum_{i \in ADV} L(X_{i}^{adv}|y_{i}) \right)
\]
where $L(X|y)$ is a loss on a single example $X$ with true class $y$;
$m$ is total number of training examples in the minibatch;
$k$ is number of adversarial examples in the minibatch and
$\lambda$ is a parameter which controls the relative weight of adversarial examples in the loss.
We used $\lambda=0.3$, $m=32$, and $k=16$.
Note that we \textit{replace} each clean example with its adversarial counterpart,
for a total minibatch size of $32$, which is a departure from previous approaches
to adversarial training. 

Fraction and weight of adversarial examples which we used in each minibatch differs
from \citet{LearningWithStrongAdversary} where authors replaced entire minibatch with adversarial examples.
However their experiments was done on smaller datasets (MNIST and CIFAR-10)
in which case adversarial training does not lead to decrease of accuracy on clean images.
We found that our approach works better for ImageNet models
(corresponding comparative experiments could be found in Appendix~\ref{app:num-adv-minibatch}).

We observed that if we fix $\epsilon$ during training then networks become robust only to
that specific value of $\epsilon$.
We therefore recommend choosing $\epsilon$ randomly, independently for each training example.
In our experiments we achieved best results when magnitudes were drawn from
a truncated normal distribution defined in interval $[0, 16]$
with underlying normal distribution $N(\mu=0, \sigma=8)$.\footnote{
  In TensorFlow this could be achieved by {\tt tf.abs(tf.truncated\_normal(shape, mean=0, stddev=8))}.
}

\section{Experiments}\label{sec:experiments}

We adversarially trained an Inception v3 model~\citep{Inception-v3} on ImageNet.
All experiments were done using synchronous distributed training on $50$ machines,
with a minibatch of $32$ examples on each machine.
We observed that the network tends to reach maximum accuracy at around $130k - 150k$ iterations.
If we continue training beyond $150k$ iterations then eventually accuracy might decrease by a fraction of a percent.
Thus we ran experiments for around $150k$ iterations and then used the obtained accuracy as the final result of the experiment.

Similar to~\citet{Inception-v3} we used RMSProp optimizer for training.
We used a learning rate of $0.045$ except where otherwise indicated.

We looked at interaction of adversarial training and other forms or regularization (dropout, label smoothing and weight decay).
By default training of Inception v3 model uses all three of them.
We noticed that disabling label smoothing and/or dropout leads to
small decrease of accuracy on clean examples (by $0.1\%$ - $0.5\%$ for top 1 accuracy)
and small increase of accuracy on adversarial examples (by $1\%$ - $1.5\%$ for top 1 accuracy).
On the other hand reducing weight decay leads to decrease of accuracy on both clean and adversarial examples.

We experimented with delaying adversarial training by $0$, $10k$, $20k$ and $40k$ iterations.
In such case we used only clean examples during the first $N$ training iterations and 
after $N$ iterations included both clean and adversarial examples in the minibatch.
We noticed that delaying adversarial training has almost no effect on accuracy on clean examples (difference in accuracy within $0.2\%$)
after sufficient number of training iterations (more than $70k$ in our case).
At the same time we noticed that larger delays of adversarial training might cause 
up to $4\%$ decline of accuracy on adversarial examples with high magnitude of adversarial perturbations.
For small $10k$ delay changes of accuracy was not statistically significant to recommend against it.
We used a delay of $10k$ because this allowed us to reuse the same partially trained model
as a starting point for many different experiments.

For evaluation we used the ImageNet validation set which contains $50,000$ images and does not intersect with the training set.

\subsection{Results of adversarial training}\label{sec:results-adv-train}

We experimented with adversarial training using several types of one-step methods.
We found that adversarial training using any type of one-step method increases robustness to all types of one-step adversarial examples
that we tested.
However there is still a gap between accuracy on clean and adversarial examples which could vary depending on
the combination of methods used for training and evaluation.

Adversarial training caused a slight (less than $1\%$) decrease of accuracy on clean examples
in our ImageNet experiments.
This differs from results of adversarial training reported
previously, where adversarial training increased accuracy on the test
set~\citep{Goodfellow-2015-adversarial,Takeru-2016,miyato2016virtual}.
One possible explanation is that adversarial training acts as a regularizer.
For datasets with few labeled examples where overfitting is the primary concern,
adversarial training reduces test error.
For datasets like ImageNet where state-of-the-art models typically have high training set
error, adding a regularizer like adversarial training can increase training set error
more than it decreases the gap between training and test set error.
Our results suggest that adversarial training should be employed in two scenarios:
\begin{enumerate}
  \item When a model is overfitting, and a regularizer is required.
  \item When security against adversarial examples is a concern. In this case,
        adversarial training is the method that provides the most security of
        any known defense, while losing only a small amount of accuracy.
\end{enumerate}

By comparing different one-step methods for adversarial training
we observed that the best results in terms or accuracy on test set are achieved using ``step l.l.'' or ``step rnd.'' method.
Moreover using these two methods helped the model to become robust to adversarial examples
generated by other one-step methods.
Thus for final experiments we used ``step l.l.'' adversarial method.

For brevity we omitted a detailed comparison of different one-step methods here,
but the reader can find it in Appendix~\ref{app:flavors-fast}.

\begin{table}[!h]
\caption{Top~1 and top~5 accuracies of an adversarially trained network on clean images and adversarial images with various
  test-time $\epsilon$.
Both training and evaluation were done using ``step l.l.'' method.
Adversarially training caused the baseline model to become robust to adversarial examples but lost some accuracy on
clean examples.
We therefore also trained a deeper model with two additional Inception blocks.
The deeper model benefits more from adversarial training in terms of robustness to adversarial perturbation,
and loses less accuracy on clean examples than the smaller model does.
}
\label{table:adversarial-training}
\begin{center}
\begin{tabular}{|l|l|c|c|c|c|c|}
\hline
                 &      &  Clean &  $\epsilon=2$ &  $\epsilon=4$ &  $\epsilon=8$ &  $\epsilon=16$ \\
\hline
Baseline          & top 1 &       78.4\% &           30.8\% &           27.2\% &           27.2\% &           29.5\% \\
(standard training)  & top 5 &       94.0\% &           60.0\% &           55.6\% &           55.1\% &           57.2\% \\
\hline
Adv. training     & top 1 &       77.6\% &           73.5\% &           74.0\% &           74.5\% &           73.9\% \\
                  & top 5 &       93.8\% &           91.7\% &           91.9\% &           92.0\% &           91.4\% \\                  
\hline
Deeper model          & top 1 &       78.7\% &           33.5\% &           30.0\% &           30.0\% &           31.6\% \\
(standard training)  & top 5 &       94.4\% &           63.3\% &           58.9\% &           58.1\% &           59.5\% \\
\hline
Deeper model     & top 1 &       78.1\% &           75.4\% &           75.7\% &           75.6\% &           74.4\% \\
(Adv. training)  & top 5 &       94.1\% &           92.6\% &           92.7\% &           92.5\% &           91.6\% \\                  
\hline
\end{tabular}
\end{center}
\end{table}

Results of adversarial training using ``step l.l.'' method are provided in Table~\ref{table:adversarial-training}.
As it can be seen from the table we were able to significantly increase
top-1 and top-5 accuracy on adversarial examples (up to $74\%$ and $92\%$ correspondingly)
to make it to be on par with accuracy on clean images.
However we lost about $0.8\%$ accuracy on clean examples.

We were able to slightly reduce the gap in the accuracy on clean images by slightly increasing the size of the model.
This was done by adding two additional Inception blocks to the model.
For specific details about Inception blocks refer to~\citet{Inception-v3}.

Unfortunately, training on one-step adversarial examples does not confer robustness to iterative
adversarial examples, as shown in Table~\ref{table:adversarial-training-iter-result}.

\begin{table}[!h]
\caption{Accuracy of adversarially trained network on iterative adversarial examples.
Adversarial training was done using ``step l.l.'' method.
Results were computed after $140k$ iterations of training.
Overall, we see that training on one-step adversarial examples does not confer resistance
to iterative adversarial examples.
}
\label{table:adversarial-training-iter-result}
\begin{center}
\begin{tabular}{|l|l|l|c|c|c|c|c|}
\hline
Adv. method  & Training      &       &  Clean  & $\epsilon=2$ & $\epsilon=4$ & $\epsilon=8$ & $\epsilon=16$ \\
\hline
Iter. l.l.   & Adv. training & top 1 &  77.4\% &       29.1\% &        7.5\% &        3.0\% &        1.5\% \\
             &               & top 5 &  93.9\% &       56.9\% &       21.3\% &        9.4\% &        5.5\% \\ \cline{2-8}
             & Baseline      & top 1 &  78.3\% &       23.3\% &        5.5\% &        1.8\% &        0.7\% \\
             &               & top 5 &  94.1\% &       49.3\% &       18.8\% &        7.8\% &        4.4\% \\
\hline
Iter. basic  & Adv. training & top 1 &  77.4\% &       30.0\% &       25.2\% &       23.5\% &       23.2\% \\
             &               & top 5 &  93.9\% &       44.3\% &       33.6\% &       28.4\% &       26.8\% \\ \cline{2-8}
             & Baseline      & top 1 &  78.3\% &       31.4\% &       28.1\% &       26.4\% &       25.9\% \\
             &               & top 5 &  94.1\% &       43.1\% &       34.8\% &       30.2\% &       28.8\% \\
\hline
\end{tabular}

\end{center}
\end{table}

We also tried to use iterative adversarial examples during training, however we were unable to gain any benefits out of it.
It is computationally costly and we were not able to obtain robustness to adversarial examples or to prevent
the procedure from reducing the accuracy on clean examples significantly.
It is possible that much larger models are necessary to achieve robustness to such a large class of inputs.

\subsection{Label leaking}\label{sec:label-leaking}

We discovered a {\em label leaking} effect:
when a model is trained on FGSM adversarial examples and
then evaluated using FGSM adversarial examples,
the accuracy on adversarial images becomes much higher than the accuracy on clean images
(see Table~\ref{table:label-leaking}).
This effect also occurs (but to a lesser degree) when using other one-step methods
that require the true label as input.

We say that label for specific example has been leaked if and only if
the model classifies an adversarial example correctly
when that adversarial example is generated using the true label
but misclassifies a corresponding adversarial example
that was created without using the true label.
If too many labels has been leaked then accuracy on adversarial examples
might become bigger than accuracy on clean examples which we observed on ImageNet dataset.

We believe that the effect occurs because one-step methods that use the true label
perform a very simple and predictable transformation that the model can learn to
recognize.
The adversarial example construction process thus inadvertently leaks information
about the true label into the input.
We found that the effect vanishes if we use adversarial example construction processes
that do not use the true label.
The effect also vanishes if an iterative method is used, presumably because the
output of an iterative process is more diverse and less predictable than the output
of a one-step process.

Overall due to the label leaking effect, we do not recommend to use FGSM or other methods
defined with respect to the true class label to evaluate robustness to adversarial examples;
we recommend to use other one-step methods that do not directly access the label instead.

We recommend to replace the true label with the most likely label predicted by the model.
Alternately, one can maximize the cross-entropy between the full distribution over all
predicted labels given the clean input and the distribution over all predicted labels given
the perturbed input \citep{Takeru-2016}.

\begin{table}[!h]
\caption{Effect of label leaking on adversarial examples.
When training and evaluation was done using FGSM
accuracy on adversarial examples was higher than on clean examples.
This effect was not happening when training and evaluation
was done using ``step l.l.'' method.
In both experiments training was done for $150k$ iterations
with initial learning rate $0.0225$.}
\label{table:label-leaking}
\begin{center}
\begin{tabular}{|l|l|c|c|c|c|c|}
\hline
                                      &       &   Clean &  $\epsilon=2$ &  $\epsilon=4$ &  $\epsilon=8$ &  $\epsilon=16$ \\
\hline
No label leaking,                     & top 1 &   77.3\% &        72.8\% &        73.1\% &        73.4\% &        72.0\% \\
training and eval using ``step l.l.'' & top 5 &   93.7\% &        91.1\% &        91.1\% &        91.0\% &        90.3\% \\
\hline
With label leaking,                   & top 1 &   76.6\% &        86.2\% &        87.6\% &        88.7\% &        87.0\% \\
training and eval using FGSM      & top 5 &   93.2\% &        95.9\% &        96.4\% &        96.9\% &        96.4\% \\
\hline
\end{tabular}
\end{center}
\end{table}

We revisited the adversarially trained MNIST classifier from \citet{Goodfellow-2015-adversarial}
and found that it too leaks labels.
The most labels are leaked with $\epsilon = 0.3$ on MNIST data in $[0, 1]$.
With that $\epsilon$, the model leaks 79 labels on the test set of 10,000 examples.
However, the amount of label leaking is small compared to the amount of error caused
by adversarial examples.
The error rate on adversarial examples exceeds the error rate on
clean examples for $\epsilon \in \{ .05, .1, .25, .3, .4, .45, .5\}$.
This explains why the label leaking effect was not noticed earlier.

\subsection{Influence of model capacity on adversarial robustness}\label{sec:model-capacity}

\begin{figure}
  \captionsetup[subfigure]{labelformat=empty}
  \centering
  \begin{subfigure}[b]{0.49\textwidth}
    \includegraphics[width=\textwidth]{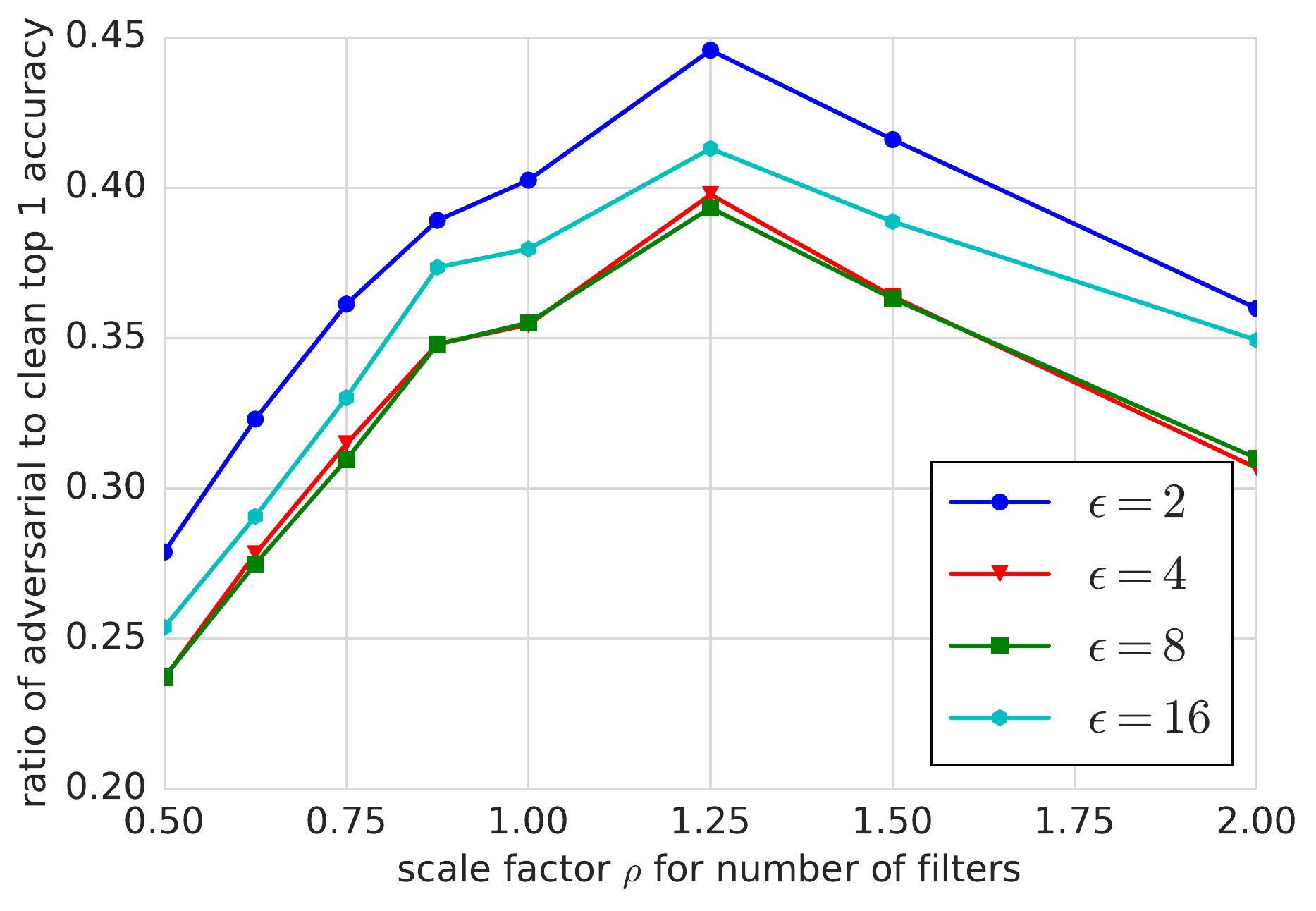}
    \caption{No adversarial training, ``step l.l.'' adv. examples}
  \end{subfigure}
  \begin{subfigure}[b]{0.49\textwidth}
    \includegraphics[width=\textwidth]{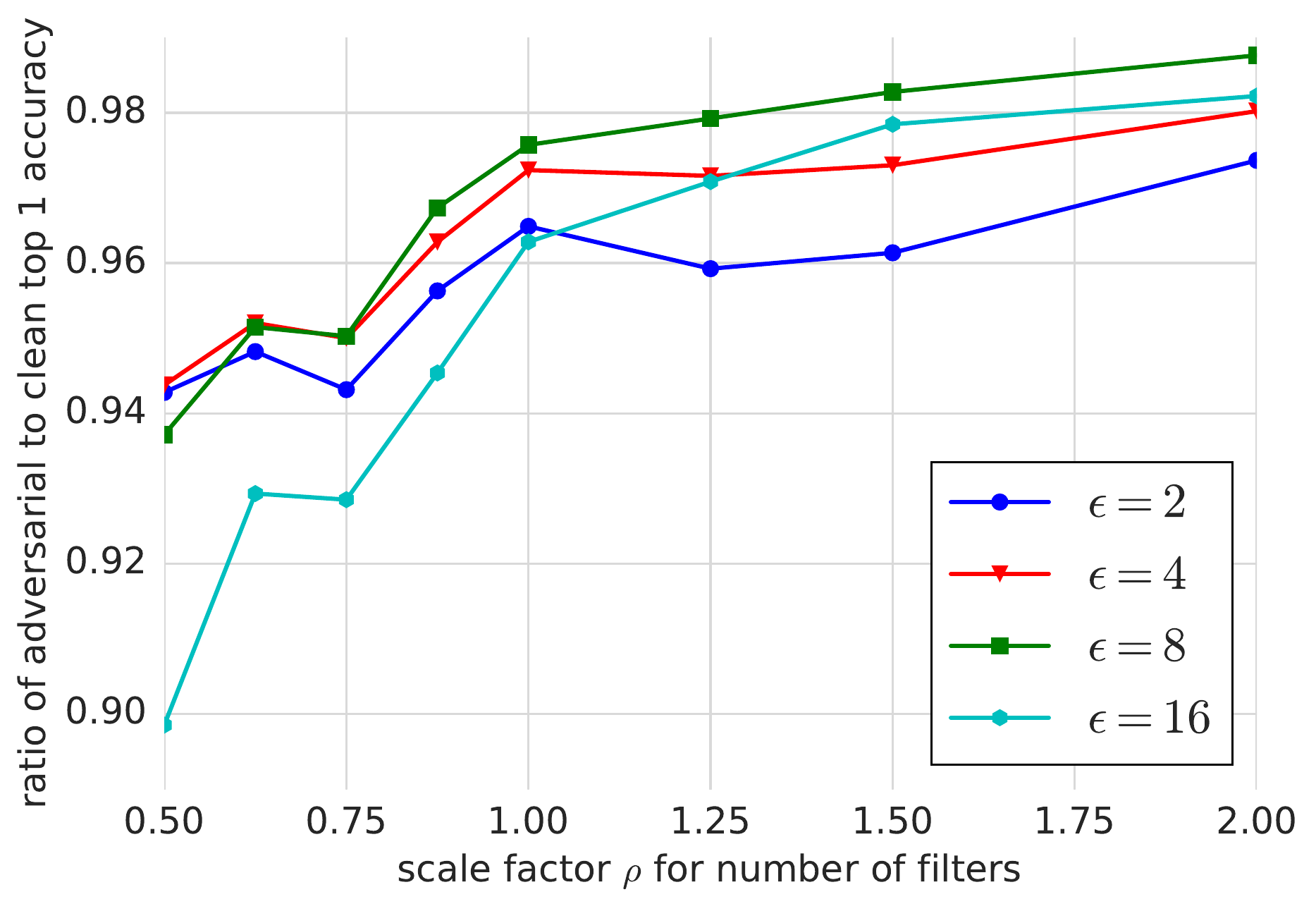}
    \caption{With adversarial training, ``step l.l.'' adv. examples}
  \end{subfigure}
  \begin{subfigure}[b]{0.49\textwidth}
    \includegraphics[width=\textwidth]{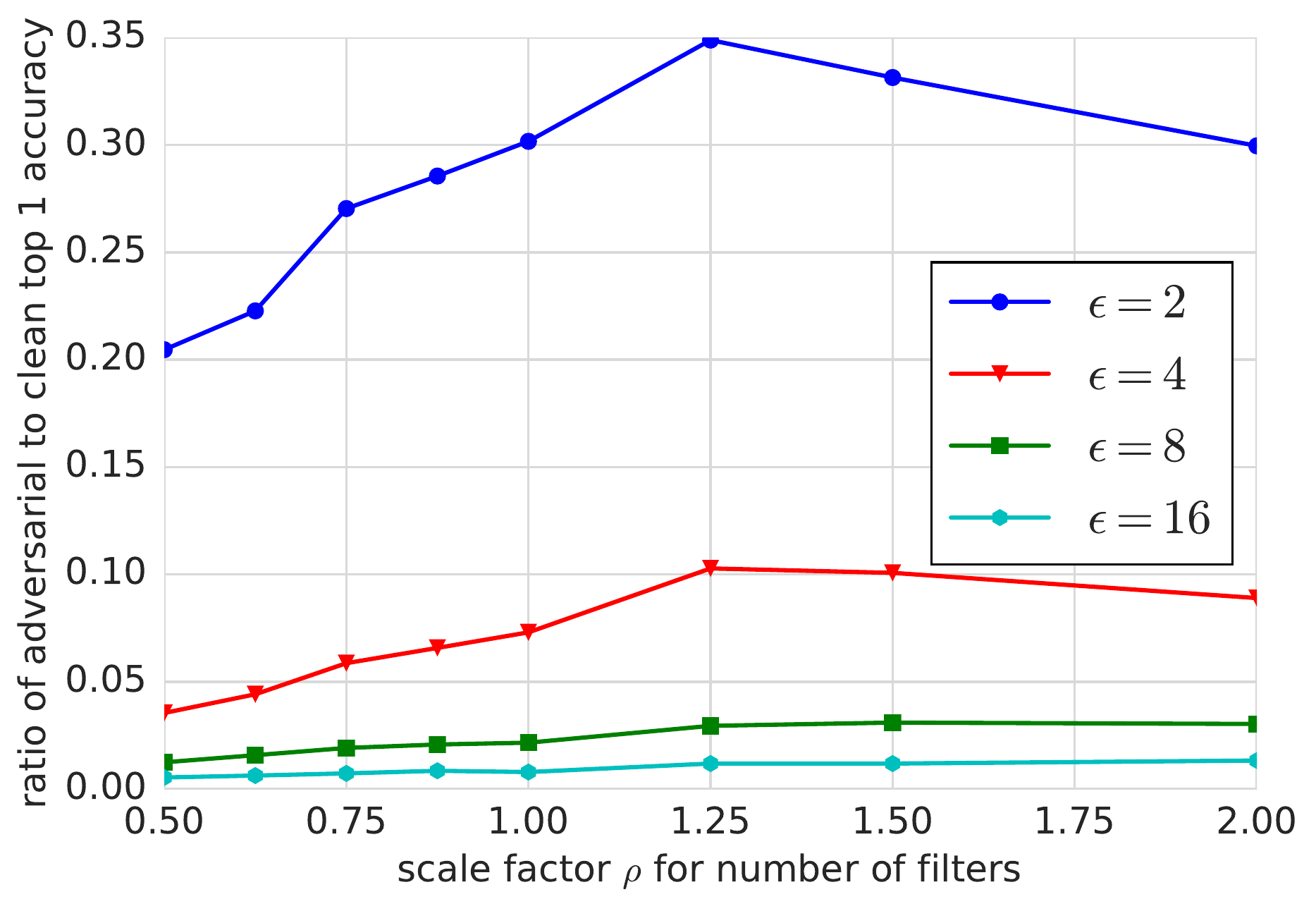}
    \caption{No adversarial training, ``iter. l.l.'' adv. examples}
  \end{subfigure}
  \begin{subfigure}[b]{0.49\textwidth}
    \includegraphics[width=\textwidth]{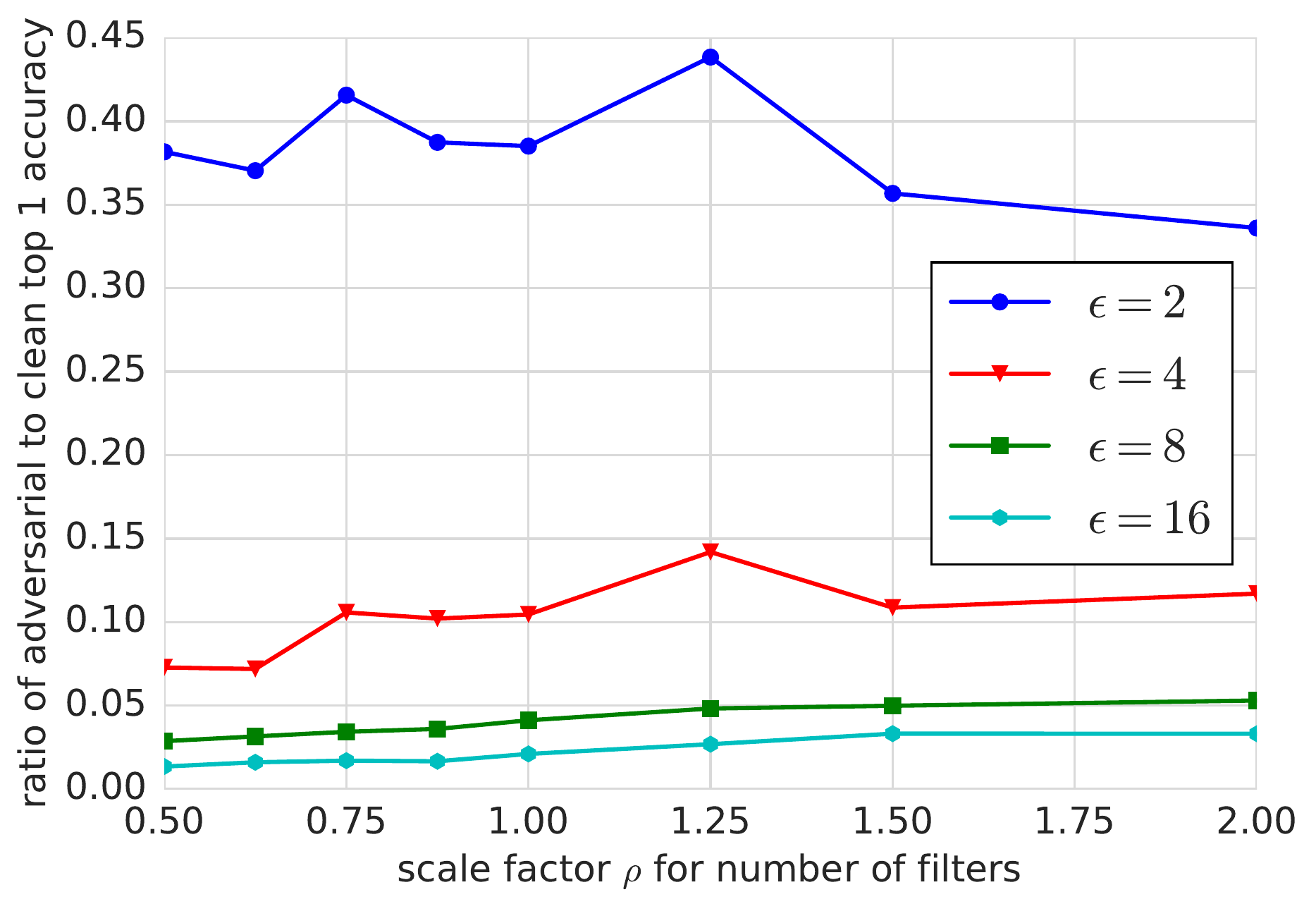}
    \caption{With adversarial training, ``iter. l.l.'' adv. examples}
  \end{subfigure}
  \begin{subfigure}[b]{0.49\textwidth}
    \includegraphics[width=\textwidth]{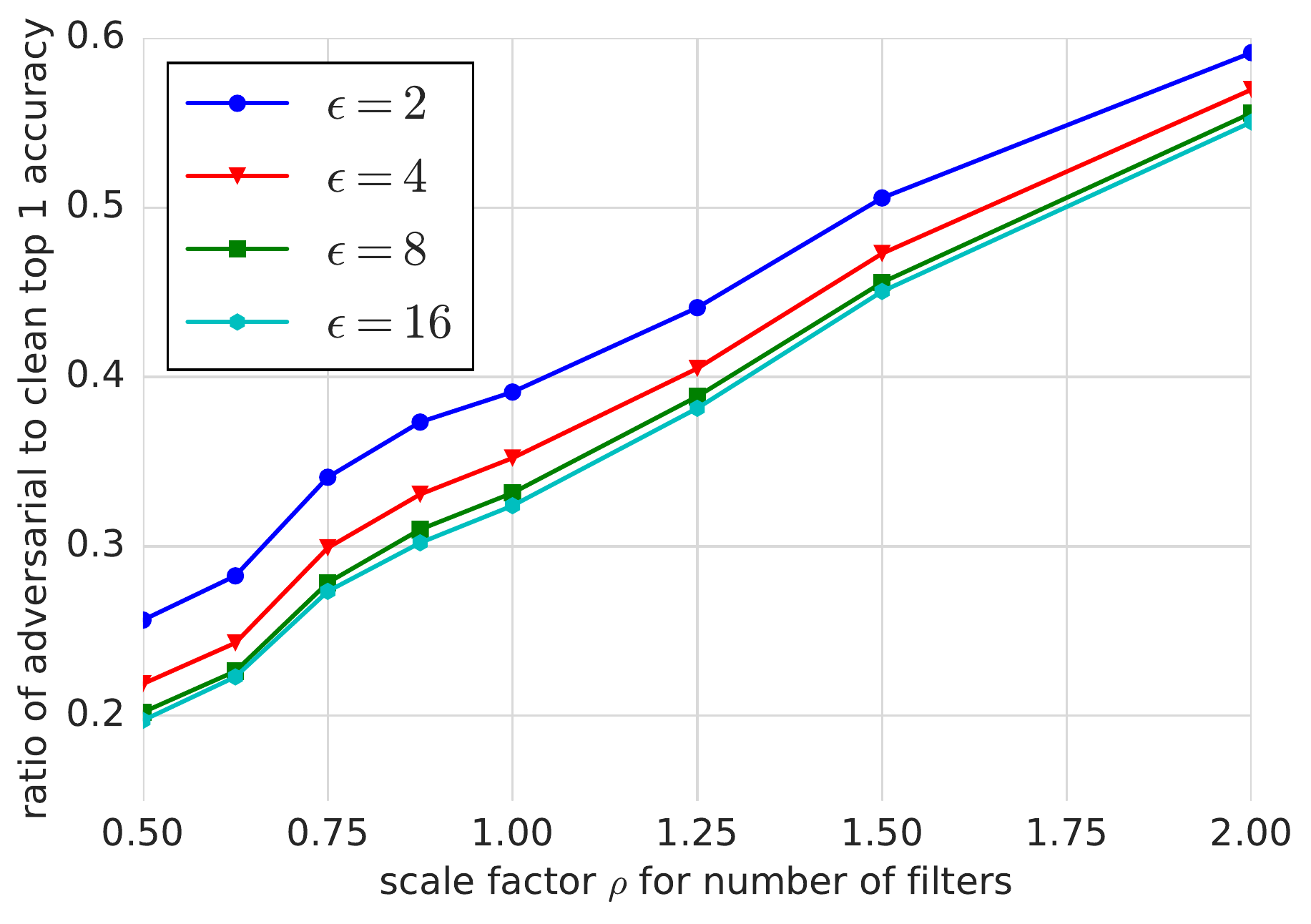}
    \caption{No adversarial training, ``basic iter.'' adv. examples}
  \end{subfigure}
  \begin{subfigure}[b]{0.49\textwidth}
    \includegraphics[width=\textwidth]{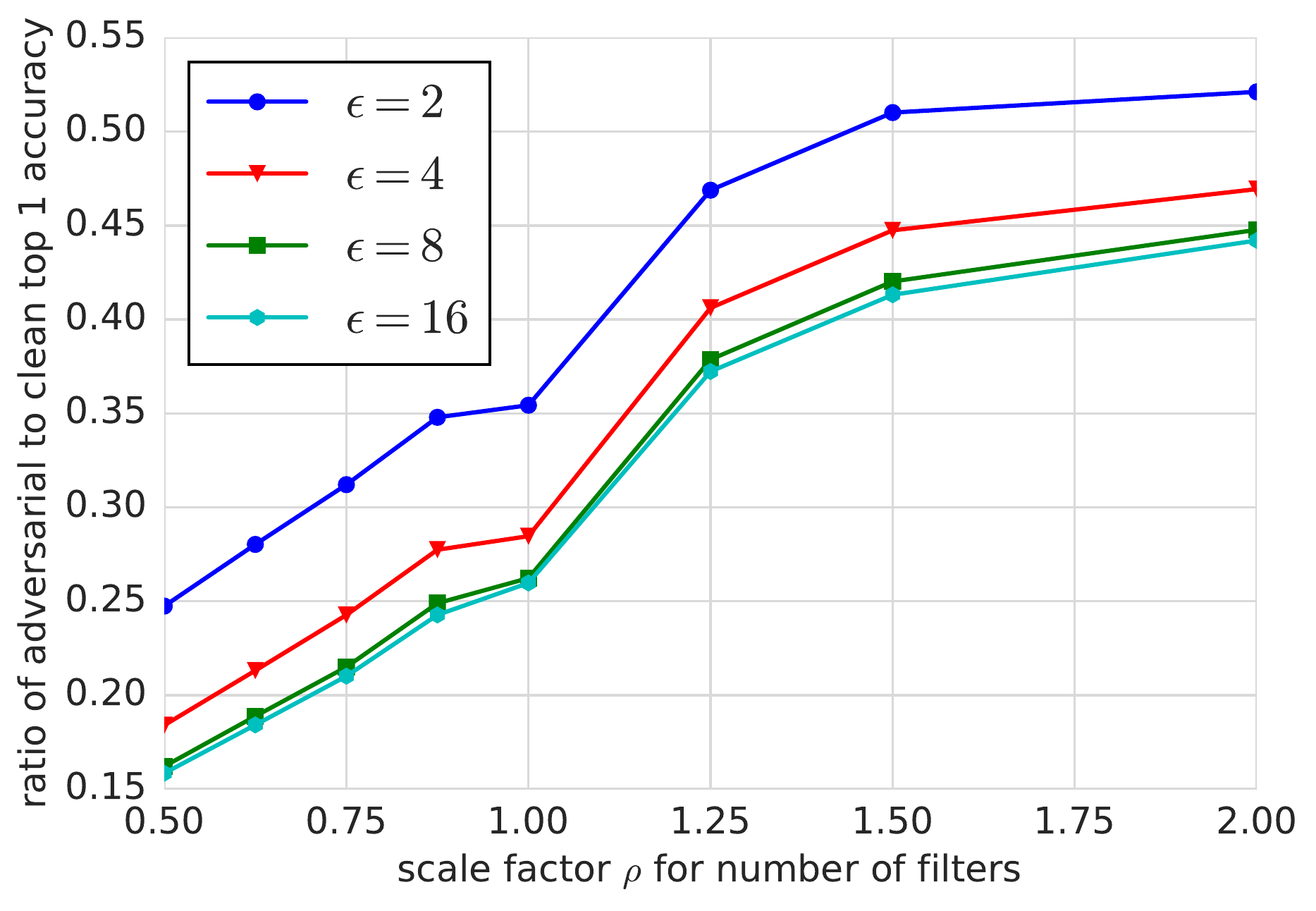}
    \caption{With adversarial training, ``basic iter.'' adv. examples}
  \end{subfigure}
  \caption{Influence of size of the model on top~1 classification accuracy of various adversarial examples.
  Left column~--- base model without adversarial training,
  right column~--- model with adversarial training using ``step l.l.'' method.
  Top row~--- results on ``step l.l.'' adversarial images,
  middle row~--- results on ``iter. l.l.'' adversarial images,
  bottom row~--- results on ``basic iter.'' adversarial images.
  See text of Section~\ref{sec:model-capacity} for explanation of meaning of horizontal and vertical axes.
  }\label{fig:capacity-influence}
\end{figure}

We studied how the size of the model (in terms of number of parameters) could affect robustness to adversarial examples.
We picked Inception v3 as a base model and varied its size by changing the number of filters in each convolution.

For each experiment we picked a scale factor $\rho$ and multiplied the number of filters in each convolution by $\rho$.
In other words $\rho=1$ means unchanged Inception v3, $\rho=0.5$ means Inception with half of the usual number of filters in convolutions, etc \ldots
For each chosen $\rho$ we trained two independent models: one with adversarial training and another without.
Then we evaluated accuracy on clean and adversarial examples for both trained models.
We have run these experiments for $\rho \in [0.5, 2.0]$.

In earlier experiments (Table \ref{table:adversarial-training}) we found that {\em deeper} models benefit
more from adversarial training. The increased depth changed many aspects of the model architecture.
These experiments varying $\rho$ examine the effect in a more controlled setting, where the architecture
remains constant except for the number of feature maps in each layer.

In all experiments we observed that accuracy on clean images kept increasing with increase of $\rho$,
though its increase slowed down as $\rho$ became bigger.
Thus as a measure of robustness we used the ratio of accuracy on adversarial images to accuracy on clean images
because an increase of this ratio means that the gap between accuracy on adversarial and clean images becomes smaller.
If this ratio reaches $1$ then the accuracy on adversarial images is the same as on clean ones.
For a successful adversarial example construction technique, we would never expect this ratio
to exceed $1$, since this would imply that the adversary is actually helpful.
Some defective adversarial example construction techniques, such as those suffering from label
leaking, can inadvertently produce a ratio greater than $1$.

Results with ratios of accuracy for various adversarial methods and $\epsilon$ are provided in Fig.~\ref{fig:capacity-influence}.

For models without adversarial training, we observed that there is an optimal value of $\rho$
yielding best robustness. Models that are too large or too small perform worse.
This may indicate that models become more robust to adversarial examples until they become
large enough to overfit in some respect.

For adversarially trained models, we found that robustness consistently increases with
increases in model size.
We were not able to train large enough models to find when this process ends, but we
did find that models with twice the normal size have an accuracy ratio approaching $1$
for one-step adversarial examples.
When evaluated on iterative adversarial examples, the trend toward increasing robustness
with increasing size remains but has some exceptions. Also, none of our models was
large enough to approach an accuracy ratio of $1$ in this regime.

Overall we recommend exploring increase of accuracy (along with adversarial training)
as a measure to improve robustness to adversarial examples.

\subsection{Transferability of adversarial examples}\label{sec:transfer}

\begin{table}[!h]
\caption{Transfer rate of adversarial examples generated using different adversarial methods and perturbation size $\epsilon=16$.
This is equivalent to the error rate in an attack scenario where the attacker prefilters their adversarial examples
by ensuring that they are misclassified by the source model before deploying them against the target.
Transfer rates are rounded to the nearest percent in order to fit the table on the page.
The following models were used for comparison: \textit{A} and \textit{B} are Inception v3 models with different random initializations,
\textit{C} is Inception v3 model with ELU activations instead of Relu, \textit{D} is Inception v4 model.
See also Table \ref{table:transfer-error-rate} for the absolute error rate when the attack is not prefiltered,
rather than the transfer rate of adversarial examples.
}
\label{table:transfer-rate}
\begin{center}
\setlength{\tabcolsep}{5pt}
\begin{tabular}{|l|l|cccc|cccc|cccc|}
\hline
& & \multicolumn{4}{|c|}{FGSM} & \multicolumn{4}{|c|}{basic iter.} & \multicolumn{4}{|c|}{iter l.l.} \\ \cline{3-14}
& source & \multicolumn{4}{|c|}{target model} & \multicolumn{4}{|c|}{target model} & \multicolumn{4}{|c|}{target model} \\
& model & A & B & C & D & A & B & C & D & A & B & C & D \\
\hline
top 1 & A (v3)     &    100 &    56 &  58 &  47 &         100 &    46 &  45 &  33 &           100 &    13 &  13 &   9 \\
      & B (v3)     &     58 &   100 &  59 &  51 &          41 &   100 &  40 &  30 &            15 &   100 &  13 &  10 \\
      & C (v3 ELU) &     56 &    58 & 100 &  52 &          44 &    44 & 100 &  32 &            12 &    11 & 100 &   9 \\
      & D (v4)     &     50 &    54 &  52 & 100 &          35 &    39 &  37 & 100 &            12 &    13 &  13 & 100 \\
\hline
top 5 & A (v3)     &    100 &    50 &  50 &  36 &         100 &    15 &  17 &  11 &           100 &     8 &   7 &   5 \\
      & B (v3)     &     51 &   100 &  50 &  37 &          16 &   100 &  14 &  10 &             7 &   100 &   5 &   4 \\
      & C (v3 ELU) &     44 &    45 & 100 &  37 &          16 &    18 & 100 &  13 &             6 &     6 & 100 &   4 \\
      & D (v4)     &     42 &    38 &  46 & 100 &          11 &    15 &  15 & 100 &             6 &     6 &   6 & 100 \\
\hline
\end{tabular}
\end{center}
\end{table}

\begin{figure}
  \captionsetup[subfigure]{labelformat=empty}
  \centering
  \begin{subfigure}[b]{0.49\textwidth}
    \includegraphics[width=\textwidth]{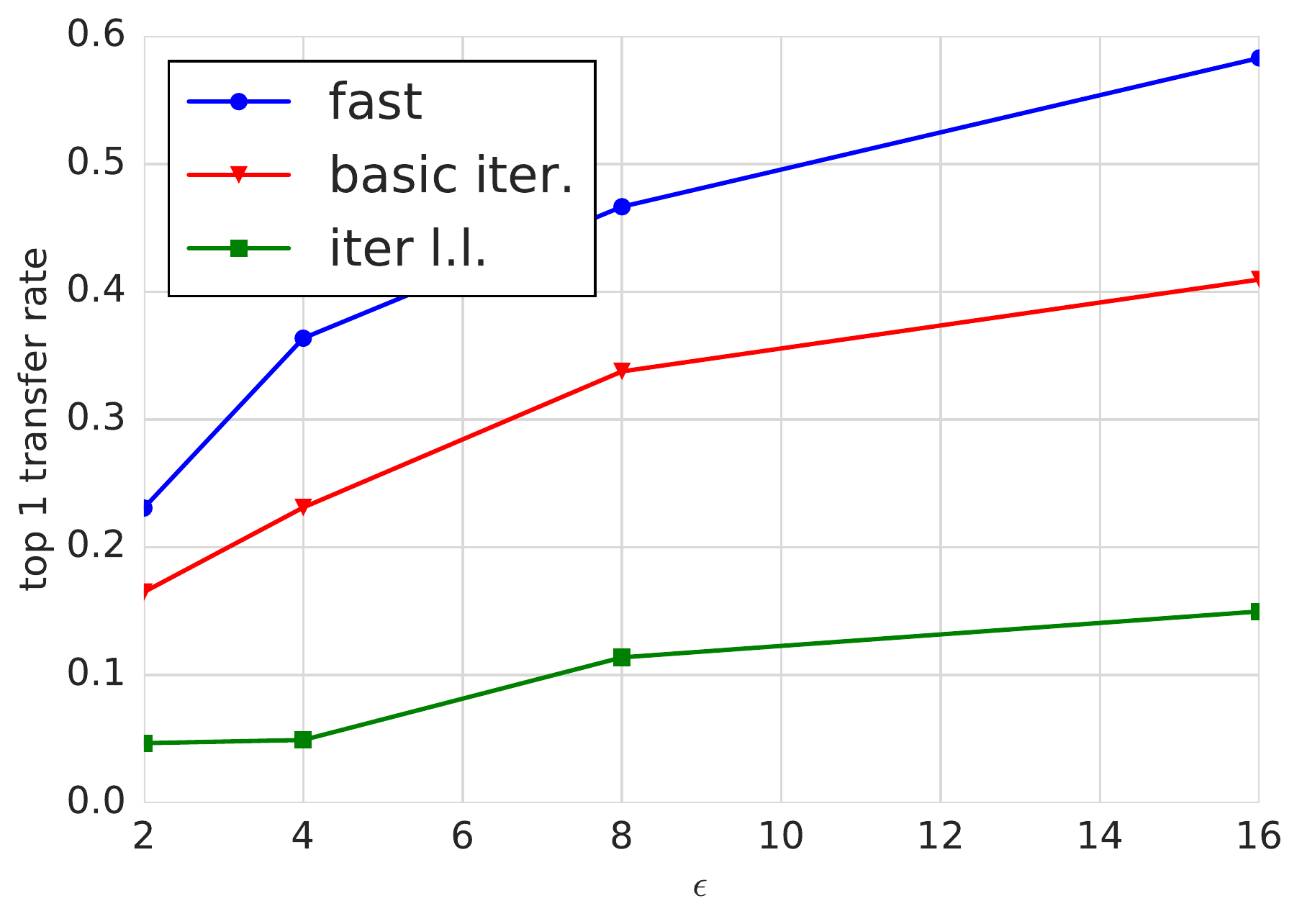}
    \caption{Top 1 transferability.}
  \end{subfigure}
  \begin{subfigure}[b]{0.49\textwidth}
    \includegraphics[width=\textwidth]{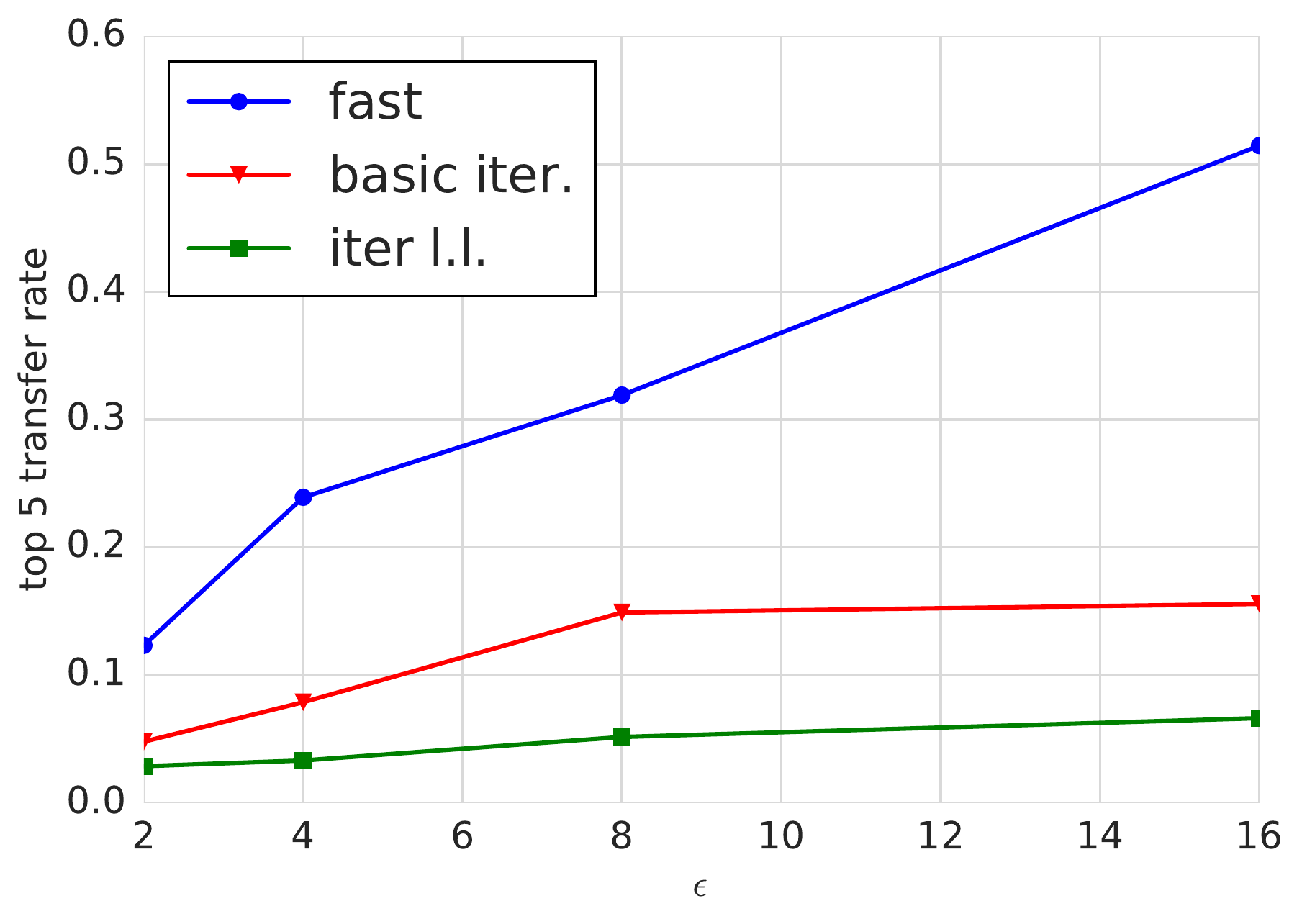}
    \caption{Top 5 transferability.}
  \end{subfigure}
  \caption{Influence of the size of adversarial perturbation on transfer rate of adversarial examples.
  Transfer rate was computed using two Inception v3 models with different random intializations.
  As could be seen from these plots, increase of $\epsilon$ leads to increase of transfer rate.
  It should be noted that transfer rate is a ratio of number of transferred adversarial examples to
  number of successful adversarial examples for source network.
  Both numerator and denominator of this ratio are increasing with increase of $\epsilon$,
  however we observed that numerator (i.e. number of transferred examples)
  is increasing much faster compared to increase of denominator.
  For example when $\epsilon$ increases from $8$ to $16$
  relative increase of denominator is less than $1\%$ for each of the considered methods,
  at the same time relative increase of numerator is more than $20\%$.}\label{fig:transferability-eps}
\end{figure}

From a security perspective, an important property of adversarial examples is
that they tend to transfer from one model to another, enabling an attacker
in the black-box scenario
to create adversarial examples for their own substitute model, then deploy
those adversarial examples to fool a target model \citep{Szegedy-ICLR2014,Goodfellow-2015-adversarial,Papernot-2016-transferability}.

We studied transferability of adversarial examples between the following models:
two copies of normal Inception v3 (with different random initializations and order or training examples),
Inception v4~\citep{Inception-v4} and
Inception v3 which uses ELU activation~\citep{EluActivation} instead of Relu\footnote{
We achieved $78.0\%$ top 1 and $94.1\%$ top 5 accuracy on Inception v3 with ELU activations,
which is comparable with accuracy of Inception v3 model with Relu activations.}.
All of these models were independently trained from scratch
until they achieved maximum accuracy.

In each experiment we fixed the source and target networks,
constructed adversarial examples from $1000$ randomly sampled clean images from the test set
using the source network and performed classification of all of them using both source and target
networks. These experiments were done independently for different adversarial methods.

We measured transferability using the following criteria.
Among $1000$ images we picked only misclassified adversarial example for the source model
(i.e. clean classified correctly, adversarial misclassified)
and measured what fraction of them were misclassified by the target model.

Transferability results for all combinations of models and $\epsilon = 16$ are provided in Table~\ref{table:transfer-rate}.
Results for various $\epsilon$ but fixed source and target model are provided in Fig.~\ref{fig:transferability-eps}.

As can be seen from the results, FGSM adversarial examples are the most transferable,
while ``iter l.l.'' are the least.
On the other hand ``iter l.l.'' method is able to fool the network in more than $99\%$ cases (top 1 accuracy),
while FGSM is the least likely to fool the network.
This suggests that there might be an inverse relationship between transferability of specific method and ability of the method to fool the network.
We haven't studied this phenomenon further, but one possible explanation could be the fact that
iterative methods tend to overfit to specific network parameters.

In addition, we observed that for each of the considered methods transfer rate is increasing with increase of $\epsilon$ (see~Fig.~\ref{fig:transferability-eps}).
Thus potential adversary performing a black-box attack have an incentive to use higher $\epsilon$ to increase the chance of success of the attack.

\section{Conclusion}\label{sec:conclusion}

In this paper we studied how to increase robustness to adversarial examples of large models (Inception~v3)
trained on large dataset (ImageNet).
We showed that adversarial training provides robustness to adversarial examples generated using one-step methods.
While adversarial training didn't help much against iterative methods we observed that
adversarial examples generated by iterative methods are less likely to be transferred between networks,
which provides indirect robustness against black box adversarial attacks.
In addition we observed that increase of model capacity could also help to increase robustness to adversarial examples
especially when used in conjunction with adversarial training.
Finally we discovered the effect of label leaking which resulted in higher accuracy on
FGSM adversarial examples compared to clean examples when the network was adversarially trained.

\bibliography{ml,adversarial_training}

\begin{thebibliography}{17}
\providecommand{\natexlab}[1]{#1}
\providecommand{\url}[1]{\texttt{#1}}
\expandafter\ifx\csname urlstyle\endcsname\relax
  \providecommand{\doi}[1]{doi: #1}\else
  \providecommand{\doi}{doi: \begingroup \urlstyle{rm}\Url}\fi

\bibitem[Biggio et~al.(2013)Biggio, Corona, Maiorca, Nelson, {\v{S}}rndi{\'c},
  Laskov, Giacinto, and Roli]{biggio2013evasion}
Battista Biggio, Igino Corona, Davide Maiorca, Blaine Nelson, Nedim
  {\v{S}}rndi{\'c}, Pavel Laskov, Giorgio Giacinto, and Fabio Roli.
\newblock Evasion attacks against machine learning at test time.
\newblock In \emph{Joint European Conference on Machine Learning and Knowledge
  Discovery in Databases}, pp.\  387--402. Springer, 2013.

\bibitem[Clevert et~al.(2015)Clevert, Unterthiner, and
  Hochreiter]{EluActivation}
Djork{-}Arn{\'{e}} Clevert, Thomas Unterthiner, and Sepp Hochreiter.
\newblock Fast and accurate deep network learning by exponential linear units
  (elus).
\newblock \emph{CoRR}, abs/1511.07289, 2015.
\newblock URL \url{http://arxiv.org/abs/1511.07289}.

\bibitem[Dalvi et~al.(2004)Dalvi, Domingos, Sanghai, Verma,
  et~al.]{dalvi2004adversarial}
Nilesh Dalvi, Pedro Domingos, Sumit Sanghai, Deepak Verma, et~al.
\newblock Adversarial classification.
\newblock In \emph{Proceedings of the tenth ACM SIGKDD international conference
  on Knowledge discovery and data mining}, pp.\  99--108. ACM, 2004.

\bibitem[Goodfellow et~al.(2014)Goodfellow, Shlens, and
  Szegedy]{Goodfellow-2015-adversarial}
Ian~J. Goodfellow, Jonathon Shlens, and Christian Szegedy.
\newblock Explaining and harnessing adversarial examples.
\newblock \emph{CoRR}, abs/1412.6572, 2014.
\newblock URL \url{http://arxiv.org/abs/1412.6572}.

\bibitem[Huang et~al.(2015)Huang, Xu, Schuurmans, and
  Szepesv{\'{a}}ri]{LearningWithStrongAdversary}
Ruitong Huang, Bing Xu, Dale Schuurmans, and Csaba Szepesv{\'{a}}ri.
\newblock Learning with a strong adversary.
\newblock \emph{CoRR}, abs/1511.03034, 2015.
\newblock URL \url{http://arxiv.org/abs/1511.03034}.

\bibitem[Ioffe \& Szegedy(2015)Ioffe and Szegedy]{Ioffe+Szegedy-2015}
Sergey Ioffe and Christian Szegedy.
\newblock Batch normalization: Accelerating deep network training by reducing
  internal covariate shift.
\newblock 2015.

\bibitem[Kurakin et~al.(2016)Kurakin, Goodfellow, and
  Bengio]{PhysicalAdversarialExamples}
Alex Kurakin, Ian Goodfellow, and Samy Bengio.
\newblock Adversarial examples in the physical world.
\newblock Technical report, arXiv, 2016.
\newblock URL \url{https://arxiv.org/abs/1607.02533}.

\bibitem[Miyato et~al.(2016{\natexlab{a}})Miyato, Dai, and
  Goodfellow]{miyato2016virtual}
Takeru Miyato, Andrew~M Dai, and Ian Goodfellow.
\newblock Virtual adversarial training for semi-supervised text classification.
\newblock \emph{arXiv preprint arXiv:1605.07725}, 2016{\natexlab{a}}.

\bibitem[Miyato et~al.(2016{\natexlab{b}})Miyato, Maeda, Koyama, Nakae, and
  Ishii]{Takeru-2016}
Takeru Miyato, Shin-ichi Maeda, Masanori Koyama, Ken Nakae, and Shin Ishii.
\newblock Distributional smoothing with virtual adversarial training.
\newblock In \emph{International Conference on Learning Representations
  (ICLR2016)}, April 2016{\natexlab{b}}.

\bibitem[{Papernot} et~al.(2016b){Papernot}, {McDaniel}, and
  {Goodfellow}]{Papernot-2016-transferability}
N.~{Papernot}, P.~{McDaniel}, and I.~{Goodfellow}.
\newblock {Transferability in Machine Learning: from Phenomena to Black-Box
  Attacks using Adversarial Samples}.
\newblock \emph{ArXiv e-prints}, May 2016b.
\newblock URL \url{http://arxiv.org/abs/1605.07277}.

\bibitem[Papernot et~al.(2015)Papernot, McDaniel, Wu, Jha, and
  Swami]{PapernotDistillation2015}
Nicolas Papernot, Patrick~Drew McDaniel, Xi~Wu, Somesh Jha, and Ananthram
  Swami.
\newblock Distillation as a defense to adversarial perturbations against deep
  neural networks.
\newblock \emph{CoRR}, abs/1511.04508, 2015.
\newblock URL \url{http://arxiv.org/abs/1511.04508}.

\bibitem[Papernot et~al.(2016a)Papernot, McDaniel, Goodfellow, Jha, Celik, and
  Swami]{Papernot-MGJCS16}
Nicolas Papernot, Patrick~Drew McDaniel, Ian~J. Goodfellow, Somesh Jha,
  Z.~Berkay Celik, and Ananthram Swami.
\newblock Practical black-box attacks against deep learning systems using
  adversarial examples.
\newblock \emph{CoRR}, abs/1602.02697, 2016a.
\newblock URL \url{http://arxiv.org/abs/1602.02697}.

\bibitem[Rozsa et~al.(2016)Rozsa, G{\"u}nther, and Boult]{rozsa2016accuracy}
Andras Rozsa, Manuel G{\"u}nther, and Terrance~E Boult.
\newblock Are accuracy and robustness correlated?
\newblock \emph{arXiv preprint arXiv:1610.04563}, 2016.

\bibitem[Russakovsky et~al.(2014)Russakovsky, Deng, Su, Krause, Satheesh, Ma,
  Huang, Karpathy, Khosla, Bernstein, et~al.]{russakovsky2014imagenet}
Olga Russakovsky, Jia Deng, Hao Su, Jonathan Krause, Sanjeev Satheesh, Sean Ma,
  Zhiheng Huang, Andrej Karpathy, Aditya Khosla, Michael Bernstein, et~al.
\newblock Imagenet large scale visual recognition challenge.
\newblock \emph{arXiv preprint arXiv:1409.0575}, 2014.

\bibitem[Szegedy et~al.(2014)Szegedy, Zaremba, Sutskever, Bruna, Erhan,
  Goodfellow, and Fergus]{Szegedy-ICLR2014}
Christian Szegedy, Wojciech Zaremba, Ilya Sutskever, Joan Bruna, Dumitru Erhan,
  Ian~J. Goodfellow, and Rob Fergus.
\newblock Intriguing properties of neural networks.
\newblock \emph{ICLR}, abs/1312.6199, 2014.
\newblock URL \url{http://arxiv.org/abs/1312.6199}.

\bibitem[Szegedy et~al.(2015)Szegedy, Vanhoucke, Ioffe, Shlens, and
  Wojna]{Inception-v3}
Christian Szegedy, Vincent Vanhoucke, Sergey Ioffe, Jonathon Shlens, and
  Zbigniew Wojna.
\newblock Rethinking the inception architecture for computer vision.
\newblock \emph{CoRR}, abs/1512.00567, 2015.
\newblock URL \url{http://arxiv.org/abs/1512.00567}.

\bibitem[Szegedy et~al.(2016)Szegedy, Ioffe, and Vanhoucke]{Inception-v4}
Christian Szegedy, Sergey Ioffe, and Vincent Vanhoucke.
\newblock Inception-v4, inception-resnet and the impact of residual connections
  on learning.
\newblock \emph{CoRR}, abs/1602.07261, 2016.
\newblock URL \url{http://arxiv.org/abs/1602.07261}.

\end{thebibliography}
\bibliographystyle{iclr2017_conference}

\newpage
\begin{appendices}

\section{Comparison of one-step adversarial methods}\label{app:flavors-fast}

In addition to FGSM and ``step l.l.'' methods we explored several other one-step adversarial methods
both for training and evaluation.
Generally all of these methods can be separated into two large categories.
Methods which try to maximize the loss (similar to FGSM) are in the first category.
The second category contains methods which try to maximize the probability of a specific
target class (similar to ``step l.l.'').
We also tried to use different types of random noise instead of adversarial images,
but random noise didn't help with robustness against adversarial examples.

The full list of one-step methods we tried is as follows:
\begin{itemize}
\item Methods increasing loss function $J$
  \begin{itemize}
  \item FGSM (described in details in Section~\ref{sec:fast-method}):
  $$\bm{X}^{adv} = \bm{X} + \epsilon \sign \bigl( \nabla_X J(\bm{X}, y_{true})  \bigr)$$
  \item FGSM-pred or fast method with predicted class.
  It is similar to FGSM but uses the label of the class predicted by the network
  instead of true class $y_{true}$.
  \item ``Fast entropy'' or fast method designed to maximize the entropy of the
    predicted distribution, thereby causing the model to become less certain of
    the predicted class.
  \item ``Fast grad. $L_2$'' is similar to FGSM but uses the value of gradient instead of its sign.
  The value of gradient is normalized to have unit $L_{2}$ norm:
  $$\bm{X}^{adv} = \bm{X} + \epsilon \frac{\nabla_X J(\bm{X}, y_{true})}{ \bigl\| \nabla_X J(\bm{X}, y_{true}) \bigr\|_{2} }$$
  \citet{Takeru-2016} advocate this method.
  \item ``Fast grad. $L_{\infty}$'' is similar to ``fast grad. $L_2$'' but uses $L_{\infty}$ norm for normalization.
  \end{itemize}
\item Methods increasing the probability of the selected target class
  \begin{itemize}
  \item ``Step l.l.'' is one-step towards least likely class (also described in Section~\ref{sec:step-ll-method}):
  $$\bm{X}^{adv} = \bm{X} - \epsilon \sign \bigl( \nabla_X J(\bm{X}, y_{target}) \bigr)$$
  where $y_{target} = \argmin_{y} \bigl\{ p( y \mid \bm{X} ) \bigr\}$ is least likely class prediction by the network.
  \item ``Step rnd.'' is similar to ``step l.l.'' but uses random class instead of least likely class.
  \end{itemize}
\item Random perturbations
  \begin{itemize}
  \item Sign of random perturbation. This is an attempt to construct random perturbation which
  has similar structure to perturbations generated by FGSM:
  $$\bm{X}^{adv} = \bm{X} + \epsilon \sign \bigl( \mathcal{N} \bigr)$$
  where $\mathcal{N}$ is random normal variable with zero mean and identity covariance matrix.
  \item Random truncated normal perturbation with zero mean and $0.5\epsilon$ standard deviation
  defined on $[-\epsilon, \epsilon]$ and uncorrelated pixels,
  which leads to the following formula for perturbed images:
  $$\bm{X}^{adv} = \bm{X} + \mathcal{T}$$
  where $\mathcal{T}$ is a random variable with truncated normal distribution.
  \end{itemize}
\end{itemize}

\begin{figure}[h]
  \captionsetup[subfigure]{labelformat=empty}
  \centering
  \begin{subfigure}[b]{0.49\textwidth}
    \includegraphics[width=\textwidth]{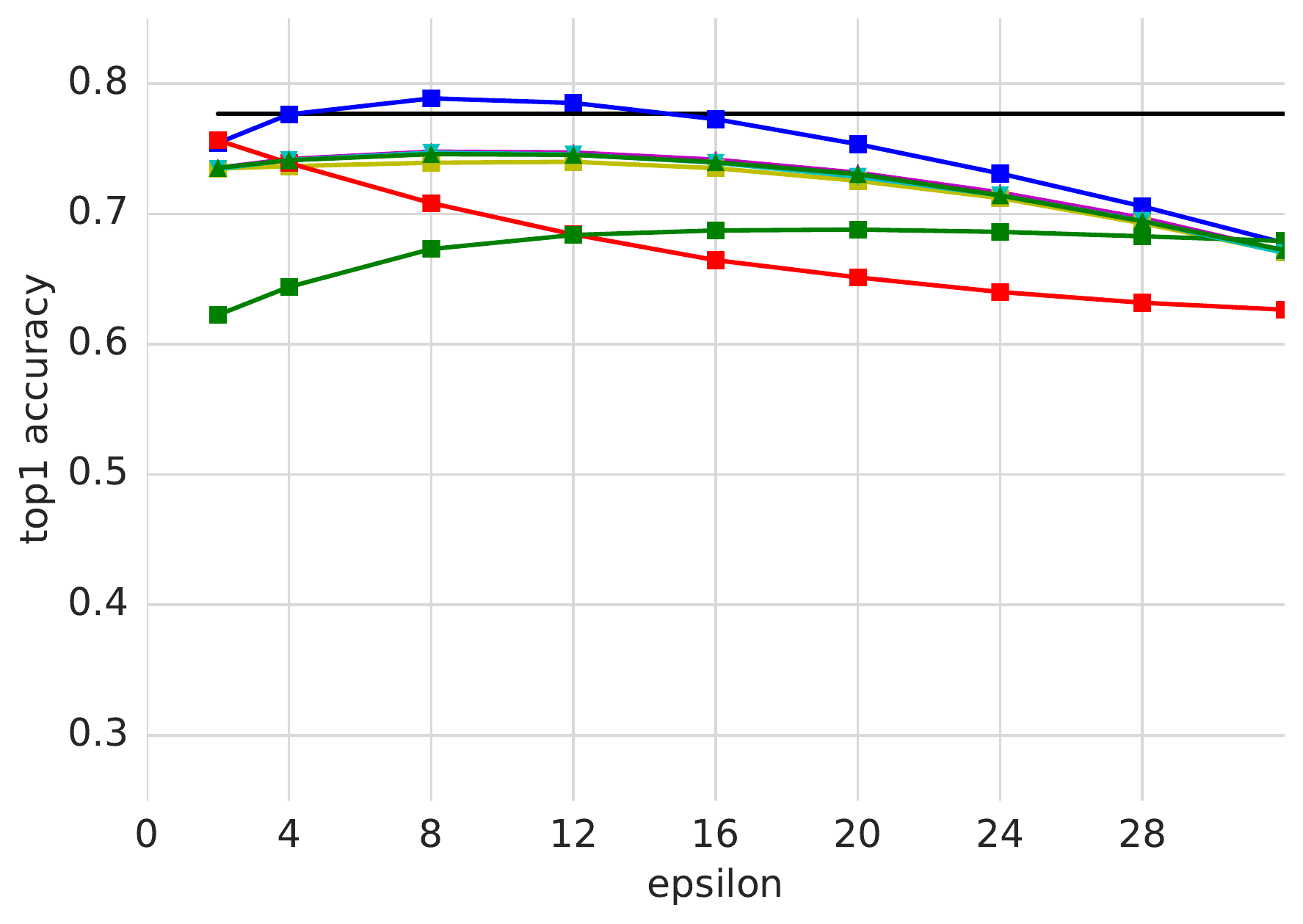}
    \caption{With adversarial training}
  \end{subfigure}
  \begin{subfigure}[b]{0.49\textwidth}
    \includegraphics[width=\textwidth]{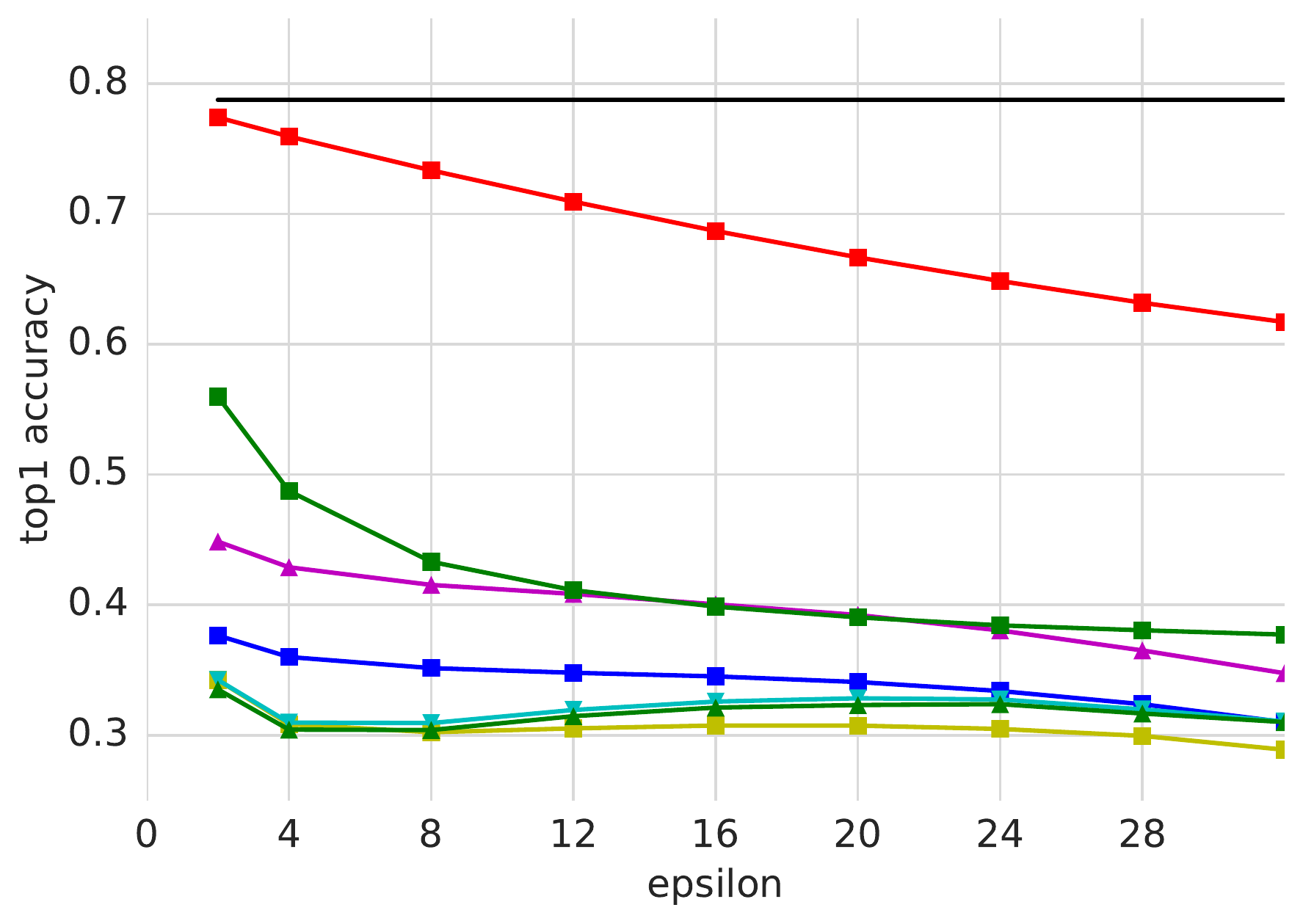}
    \caption{No adversarial training}
  \end{subfigure}
  \begin{subfigure}[b]{0.49\textwidth}
    \includegraphics[width=\textwidth]{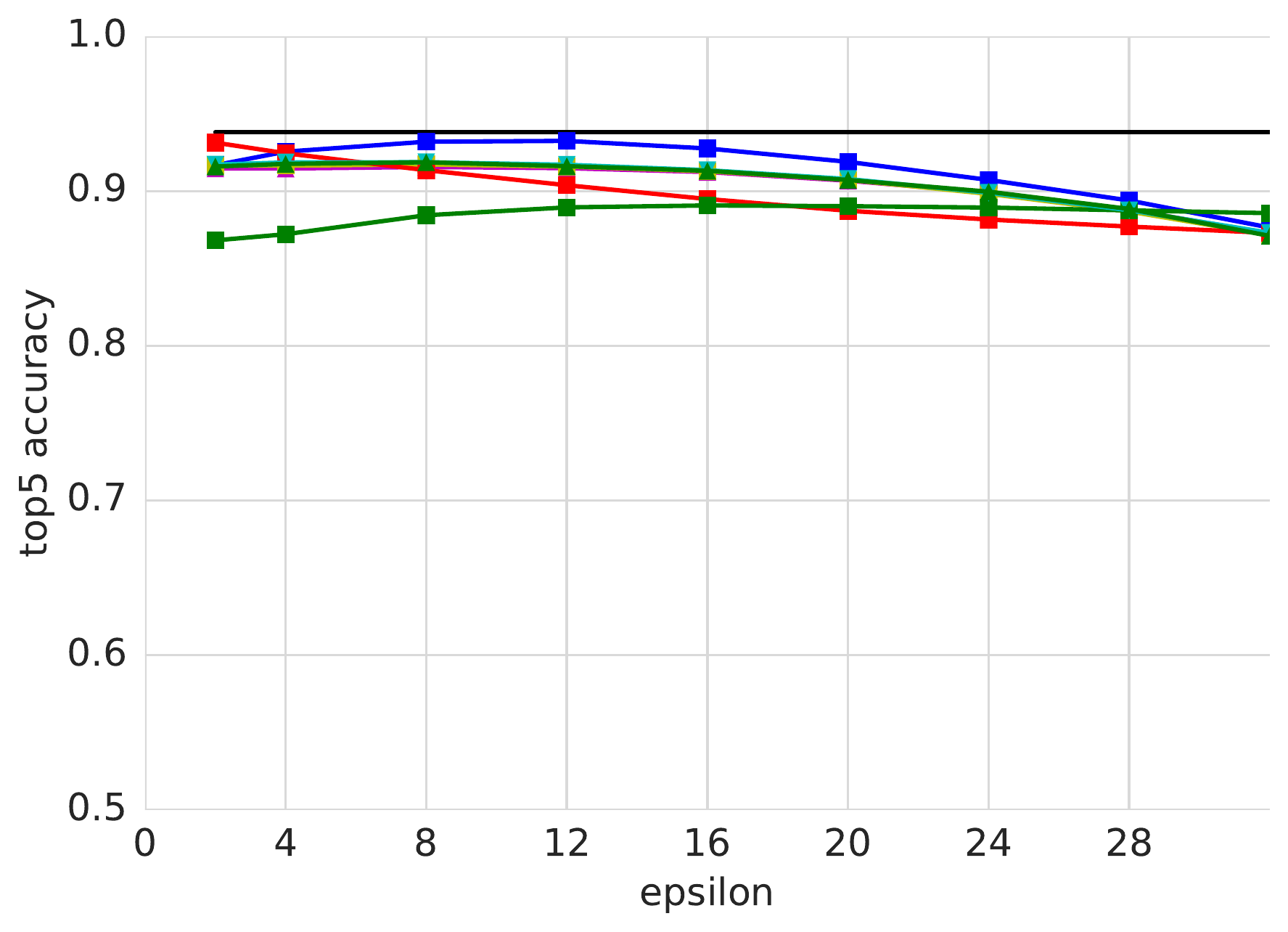}
    \caption{With adversarial training}
  \end{subfigure}
  \begin{subfigure}[b]{0.49\textwidth}
    \includegraphics[width=\textwidth]{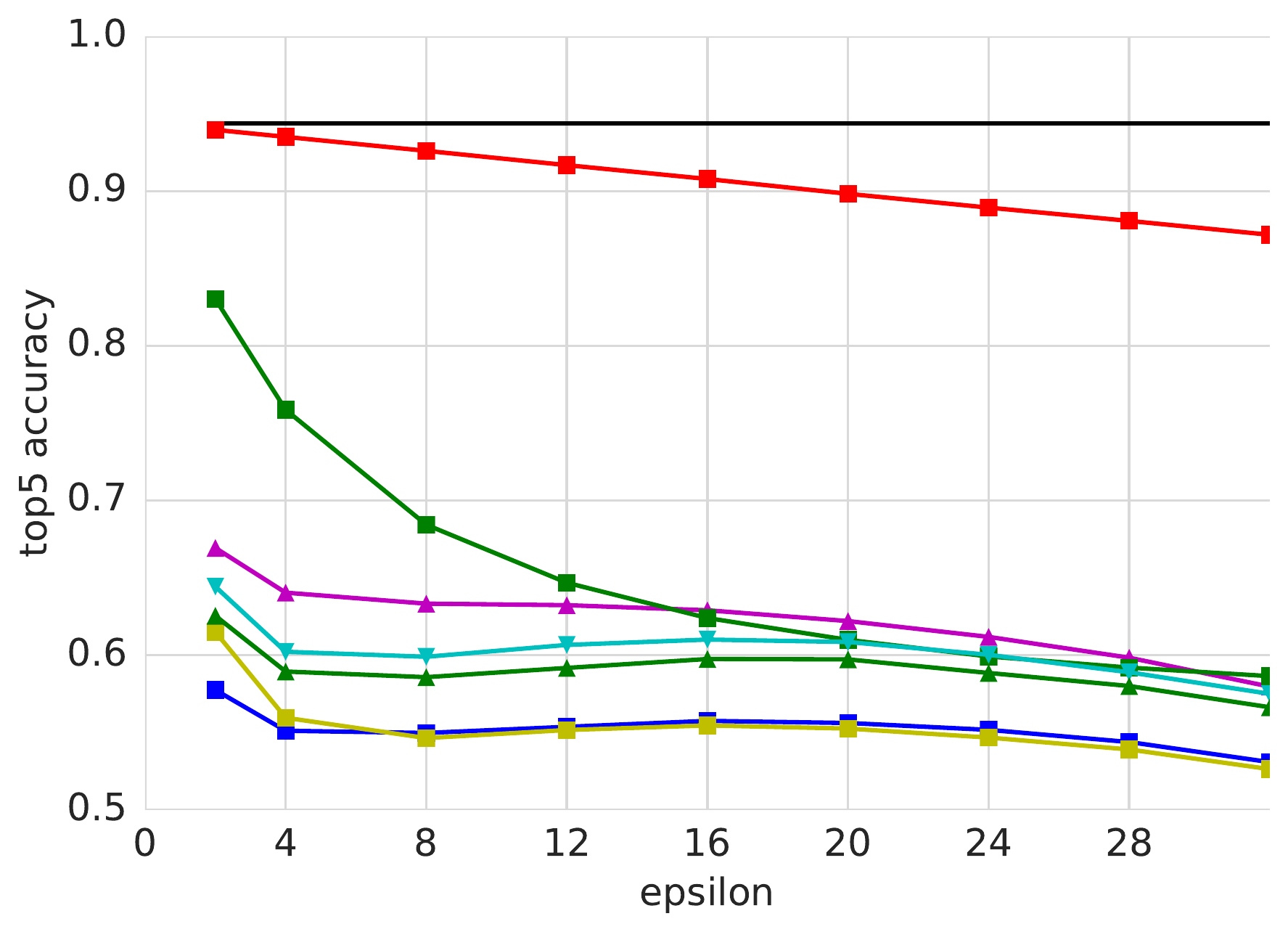}
    \caption{No adversarial training}
  \end{subfigure}
  \begin{subfigure}[b]{0.98\textwidth}
    \includegraphics[width=\textwidth]{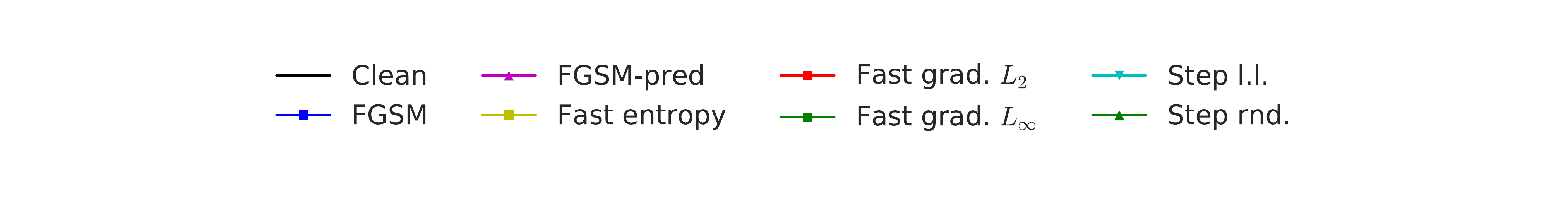}
  \end{subfigure}
  \caption{Comparison of different one-step adversarial methods during eval.
  Adversarial training was done using ``step l.l.'' method.
Some evaluation methods show increasing accuracy with increasing $\epsilon$
over part of the curve, due to the label leaking effect.
}\label{fig:eval-one-step-comparison}
\end{figure}

Overall, we observed that using only one of these single step methods during adversarial training
is sufficient to gain robustness to all of them.
Fig.~\ref{fig:eval-one-step-comparison} shows accuracy on various one-step adversarial examples
when the network was trained using only ``step l.l.'' method.

At the same time we observed that not all one-step methods are equally good for adversarial training,
as shown in Table~\ref{table:train-one-step-comparison}.
The best results (achieving both good accuracy on clean data and good accuracy on adversarial
inputs) were obtained when adversarial training was done using ``step l.l.'' or
``step rnd.'' methods.

\begin{table}[h]
\caption{Comparison of different one-step adversarial methods for adversarial training.
The evaluation was run after $90k$ training steps.
\\
*) In all cases except ``fast grad $L_2$'' and ``fast grad $L_{\infty}$''
the evaluation was done using FGSM.
For ``fast grad $L_2$'' and ``fast grad $L_{\infty}$'' the evaluation was done using ``step l.l.'' method.
In the case where both training and testing were done with FGSM,
the performance on adversarial examples is artificially high due to the label
leaking effect.
Based on this table, we recommend using ``step rnd.'' or ``step l.l.'' as the method
of generating adversarial examples at training time, in order to obtain good accuracy
on both clean and adversarial examples.
We computed 95\% confidence intervals based on the standard error of the mean
around the test error, using the fact that the test error was evaluated with 50,000
samples. Within each column, we indicate which methods are statistically tied for the
best using bold face.
}
\label{table:train-one-step-comparison}
\begin{center}
\begin{tabular}{|l|c|c|c|c|c|}
\hline
                              &   Clean &  $\epsilon=2$ &  $\epsilon=4$ &  $\epsilon=8$ &  $\epsilon=16$ \\
\hline
No adversarial training       &  {\bf 76.8\% } &       40.7\% &       39.0\% &       37.9\% &       36.7\% \\
\hline
FGSM                          &  74.9\% &      \cancel{79.3\%} &       \cancel{82.8\%} &       \cancel{85.3\%} &       \cancel{83.2\%} \\
Fast with predicted class     &  {\bf 76.4\% } &       43.2\% &       42.0\% &       40.9\% &       40.0\% \\
Fast entropy      &  {\bf 76.4\% } &       62.8\% &       61.7\% &       59.5\% &       54.8\% \\
Step rnd.                     &  {\bf 76.4\% } &       {\bf 73.0\%} &       {\bf 75.4\% } &       {\bf 76.5\%} &       {\bf 72.5\%} \\
Step l.l.                     &  {\bf 76.3\% } &       {\bf 72.9\%} &       {\bf 75.1\% } &       {\bf 76.2\%} &       {\bf 72.2\%} \\
\hline
Fast grad. $L_2$*              &  {\bf 76.8\% } &       44.0\% &       33.2\% &       26.4\% &       22.5\% \\
Fast grad. $L_{\infty}$*       &  75.6\% &       52.2\% &       39.7\% &       30.9\% &       25.0\% \\
\hline
Sign of random perturbation   &  {\bf 76.5\% } &       38.8\% &       36.6\% &       35.0\% &       32.7\% \\
Random normal perturbation    &  {\bf 76.6\% } &       38.3\% &       36.0\% &       34.4\% &       31.8\% \\
\hline
\end{tabular}
\end{center}
\end{table}

\section{Additional results with size of the model}\label{app:model-size}

\begin{figure}[h]
  \captionsetup[subfigure]{labelformat=empty}
  \centering
  \begin{subfigure}[b]{0.49\textwidth}
    \includegraphics[width=\textwidth]{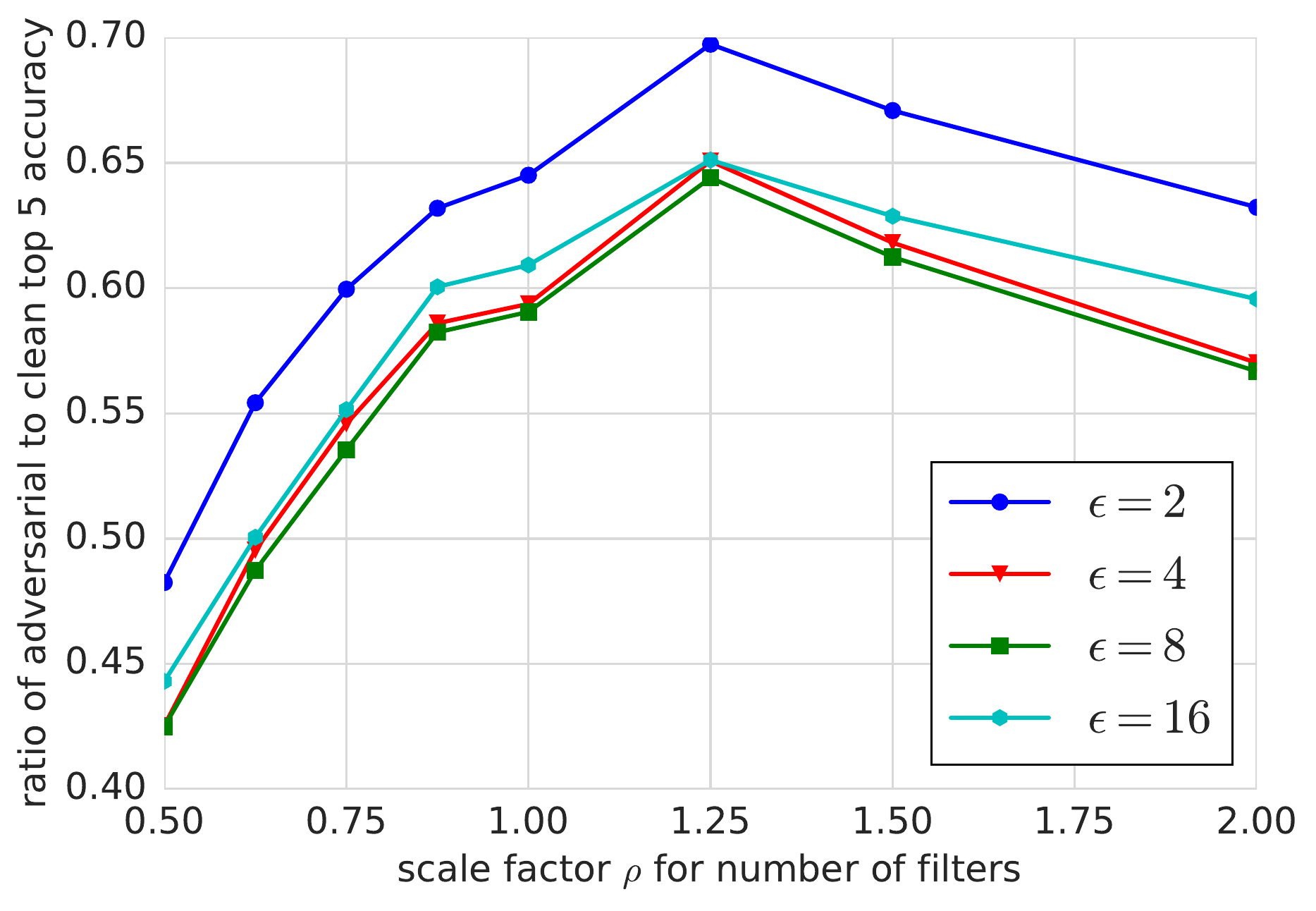}
    \caption{No adversarial training, ``step l.l.'' adv. examples}
  \end{subfigure}
  \begin{subfigure}[b]{0.49\textwidth}
    \includegraphics[width=\textwidth]{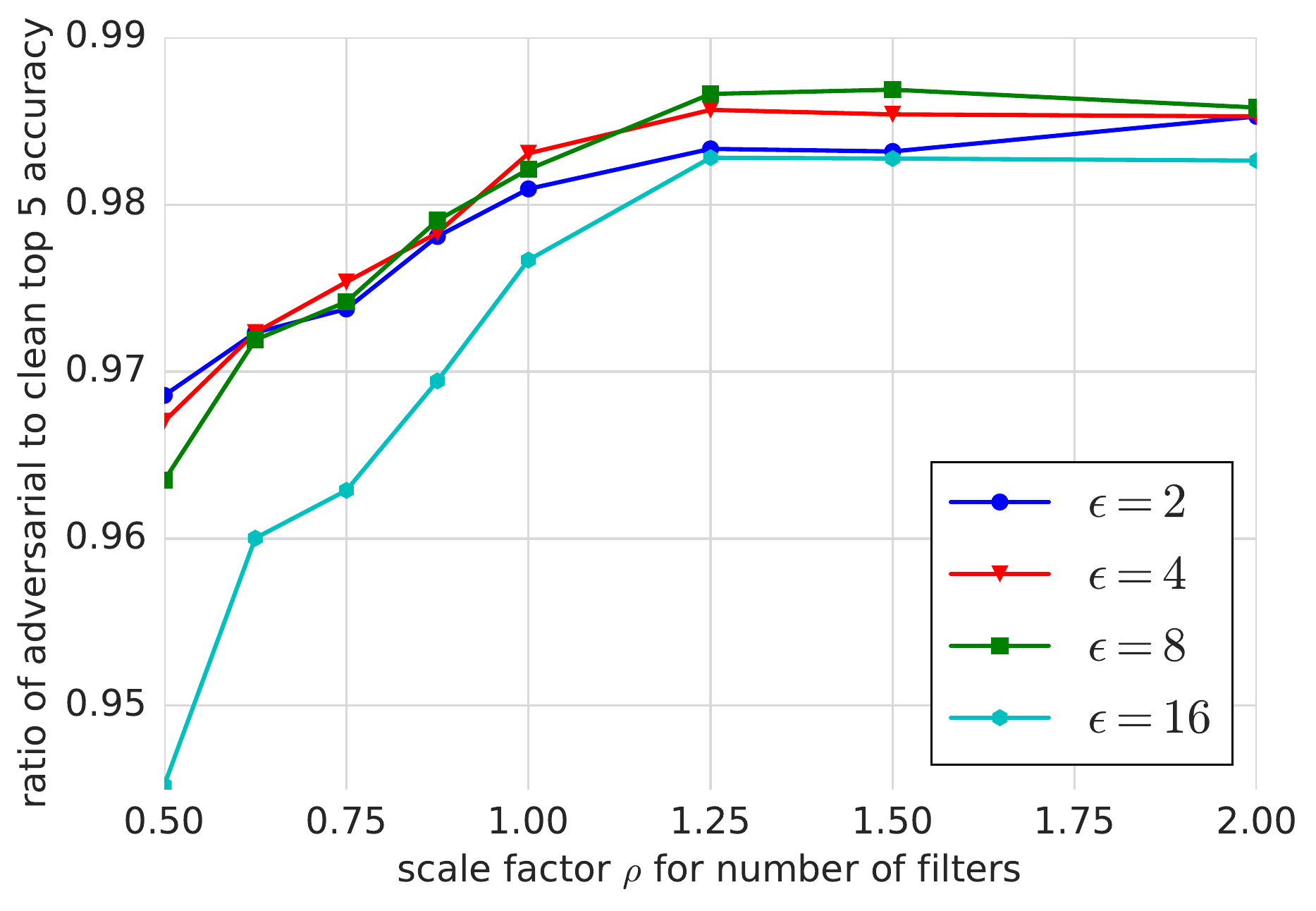}
    \caption{With adversarial training, ``step l.l.'' adv. examples}
  \end{subfigure}
  \begin{subfigure}[b]{0.49\textwidth}
    \includegraphics[width=\textwidth]{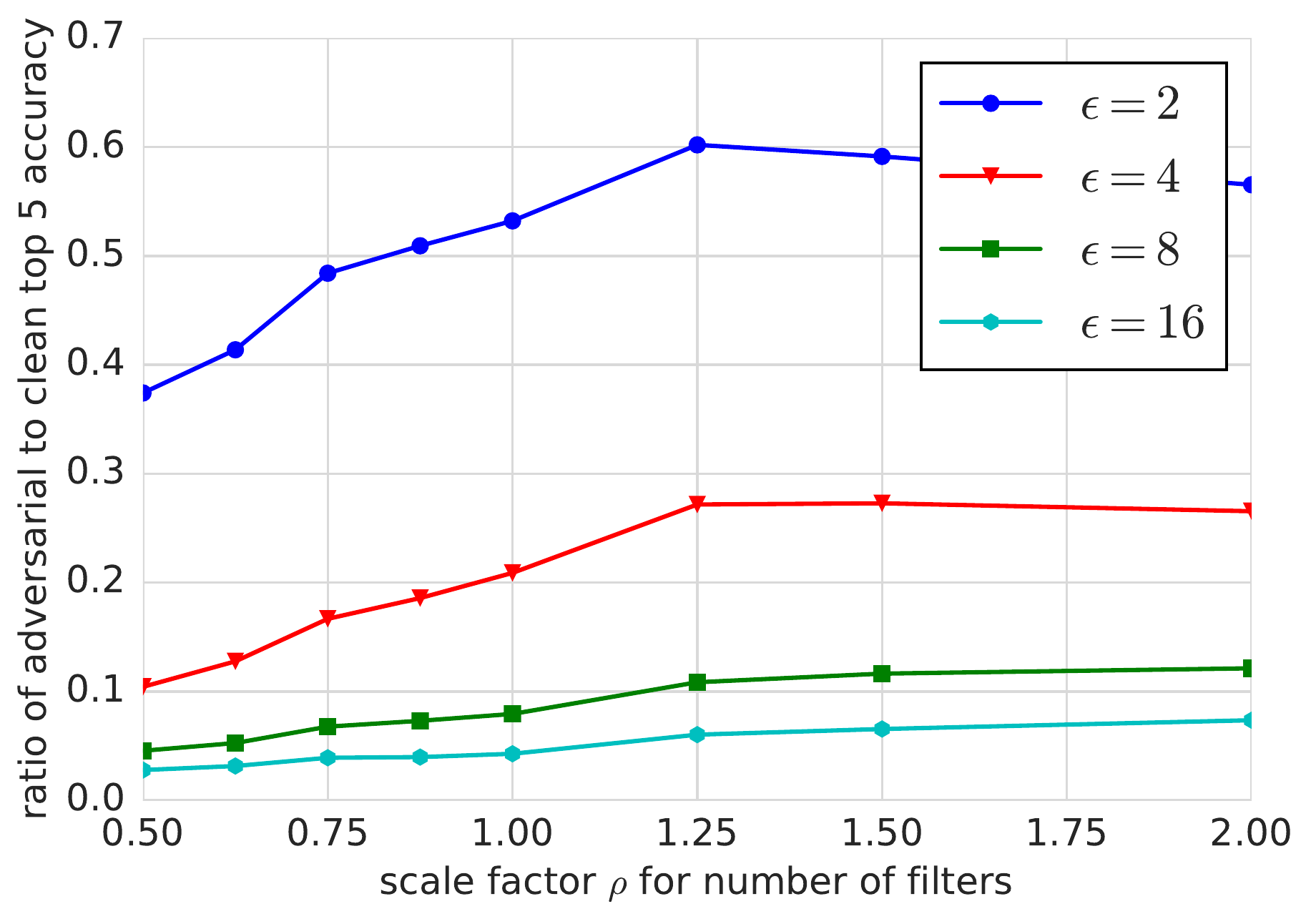}
    \caption{No adversarial training, ``iter. l.l.'' adv. examples}
  \end{subfigure}
  \begin{subfigure}[b]{0.49\textwidth}
    \includegraphics[width=\textwidth]{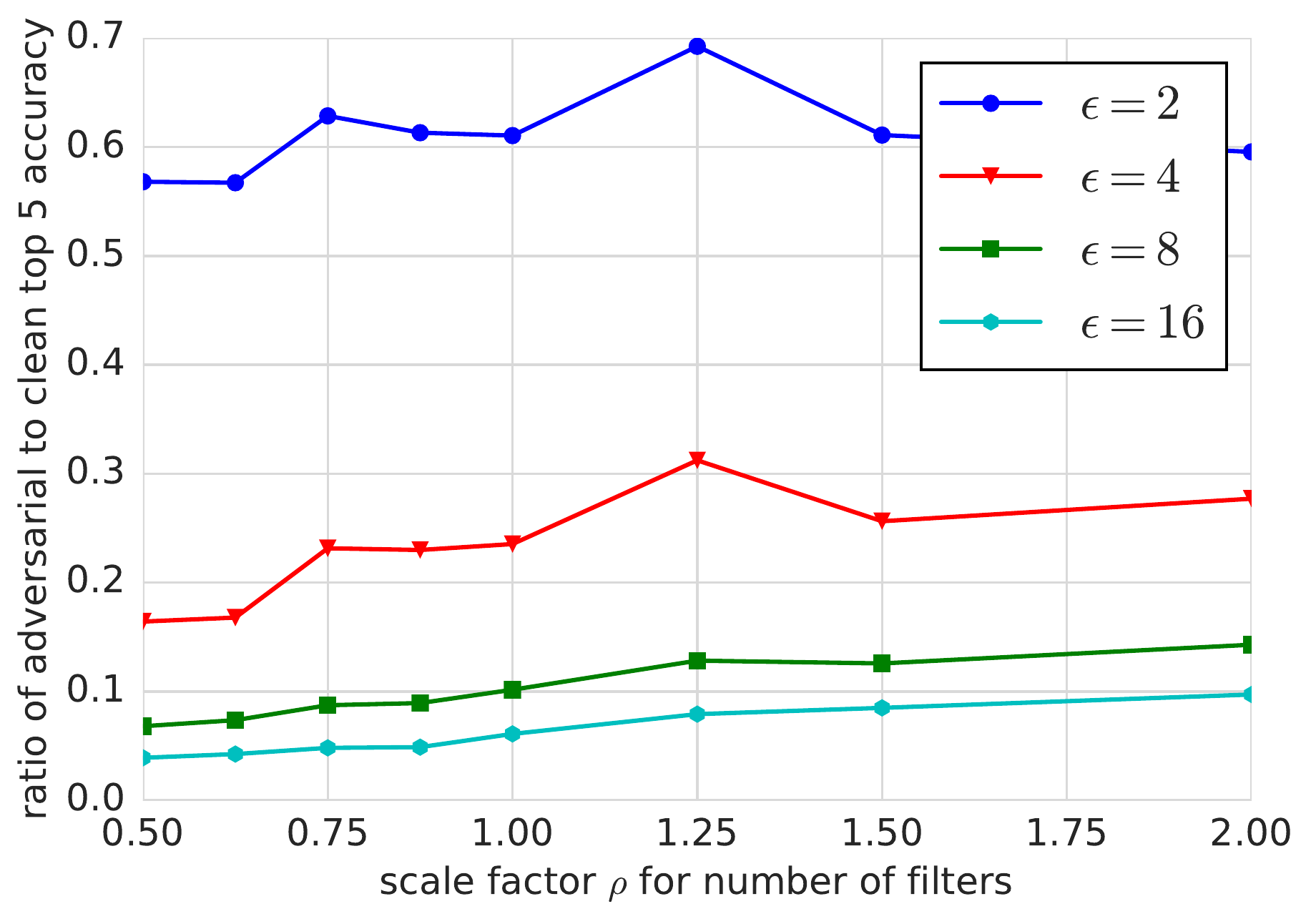}
    \caption{With adversarial training, ``iter. l.l.'' adv. examples}
  \end{subfigure}
  \begin{subfigure}[b]{0.49\textwidth}
    \includegraphics[width=\textwidth]{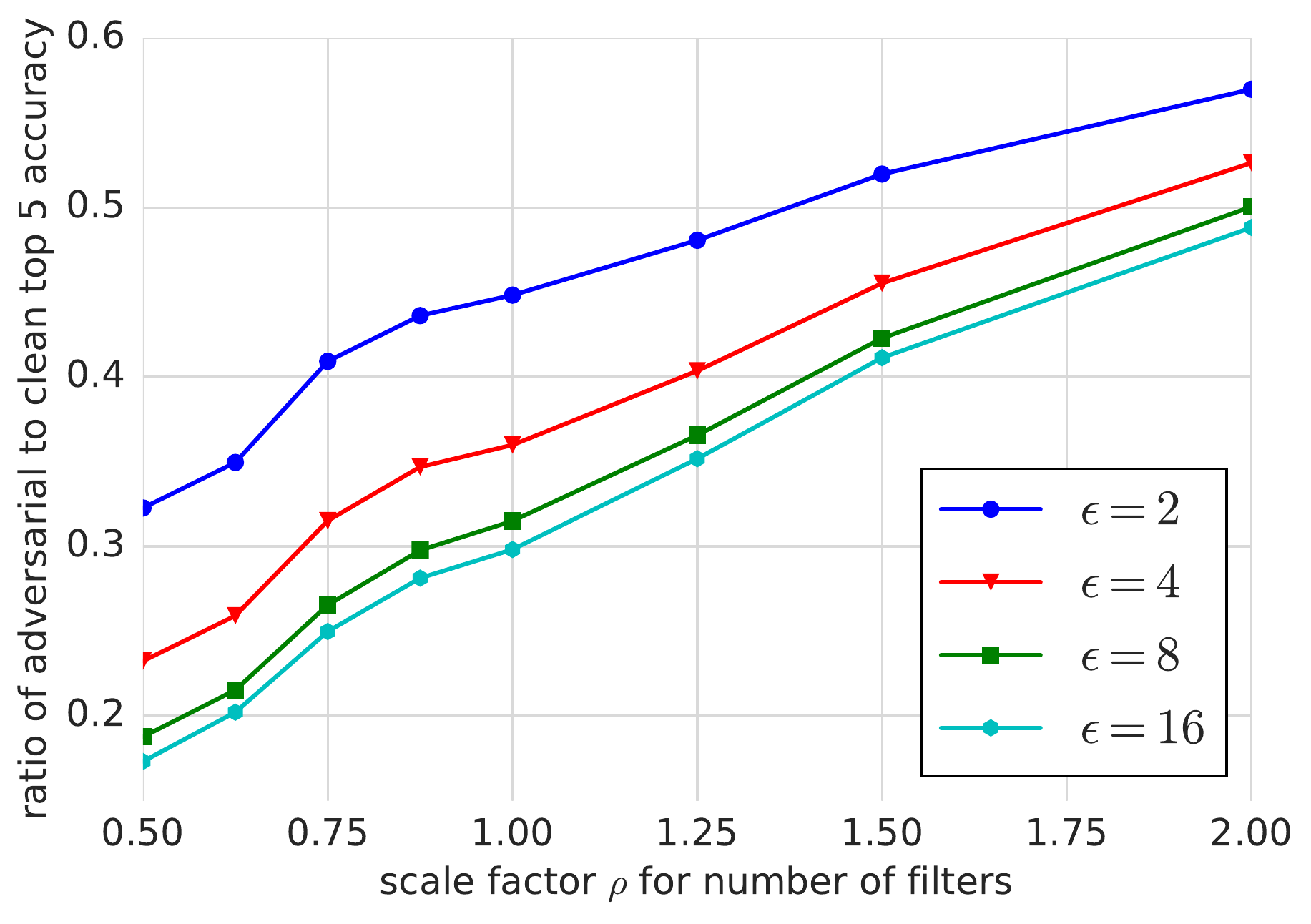}
    \caption{No adversarial training, ``basic iter.'' adv. examples}
  \end{subfigure}
  \begin{subfigure}[b]{0.49\textwidth}
    \includegraphics[width=\textwidth]{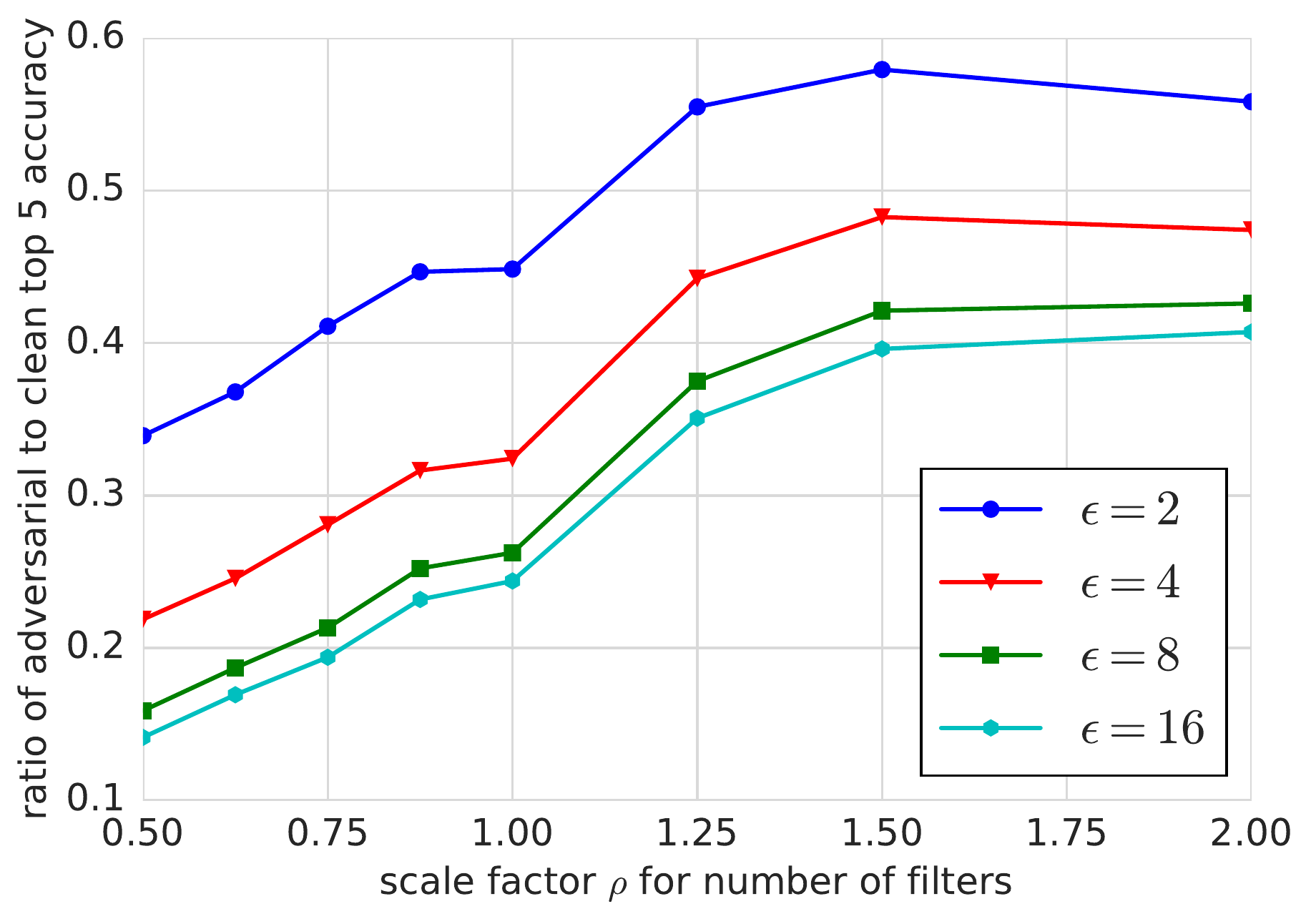}
    \caption{With adversarial training, ``basic iter.'' adv. examples}
  \end{subfigure}
  \caption{Influence of size of the model on top~5 classification accuracy of various adversarial examples.
  For a detailed explanation see Section~\ref{sec:model-capacity} and Figure~\ref{fig:capacity-influence}.
  }\label{fig:capacity-influence-top5}
\end{figure}

Section~\ref{sec:model-capacity} contains details regarding the influence of size of the model on robustness
to adversarial examples.
Here we provide additional Figure~\ref{fig:capacity-influence-top5} which shows robustness
calculated using top~5 accuracy.
Generally it exhibits the same properties as the corresponding plots for top~1 accuracy.

\section{Additional results on transferability}\label{app:transferability}

Section~\ref{sec:transfer} contains results with transfer rate of various adversarial examples between models.
In addition to transfer rate computed only on misclassified adversarial examples
it is also interesting to observe the error rate of all candidate adversarial examples generated for one model
and classified by other model.

This result might be interesting because it models the following attack.
Instead of trying to pick ``good'' adversarial images
an adversary tries to modify all available images in order to
get as much misclassified images as possible.

\begin{figure}
  \captionsetup[subfigure]{labelformat=empty}
  \centering
  \begin{subfigure}[b]{0.49\textwidth}
    \includegraphics[width=\textwidth]{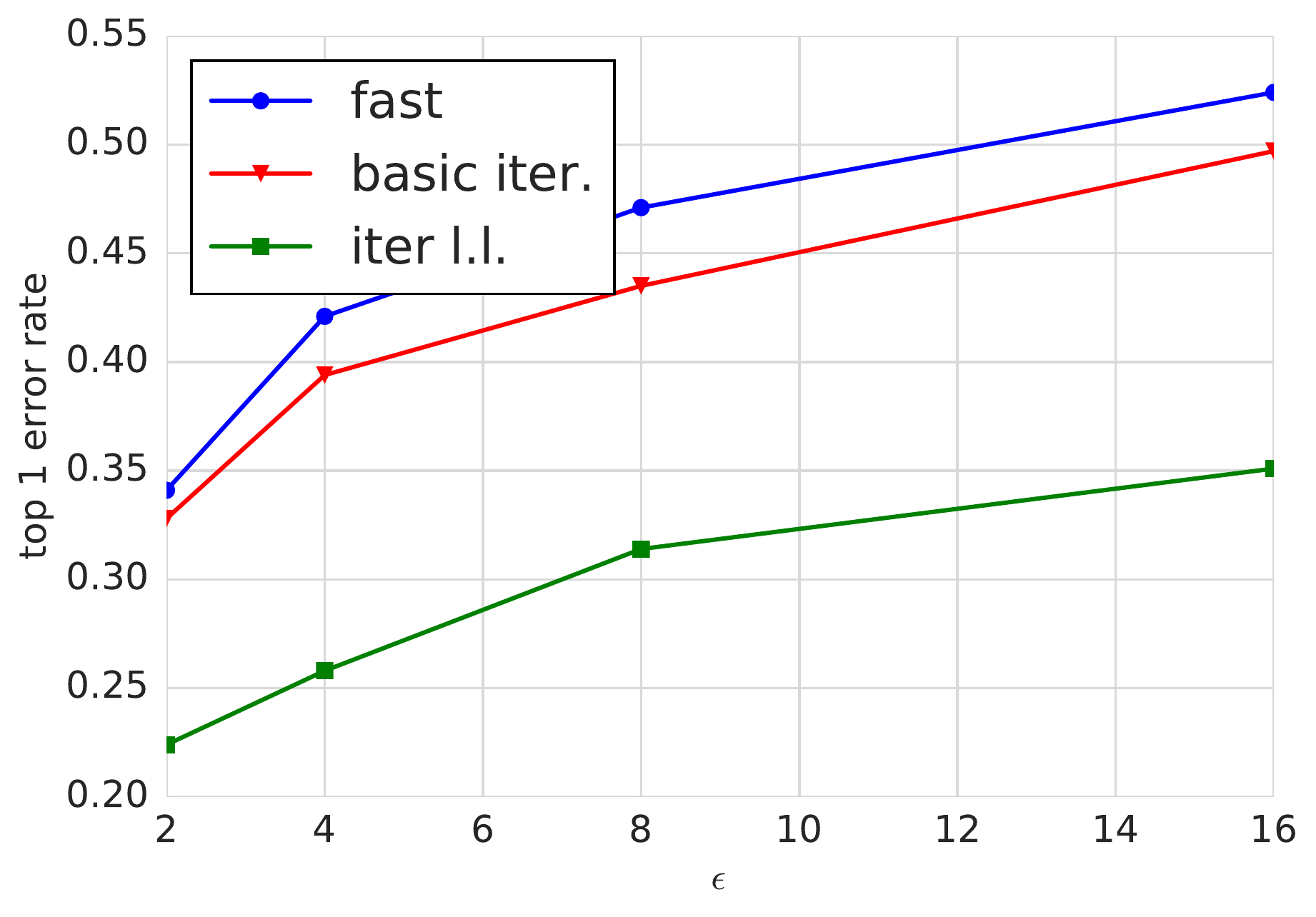}
    \caption{Top 1 error rate.}
  \end{subfigure}
  \begin{subfigure}[b]{0.49\textwidth}
    \includegraphics[width=\textwidth]{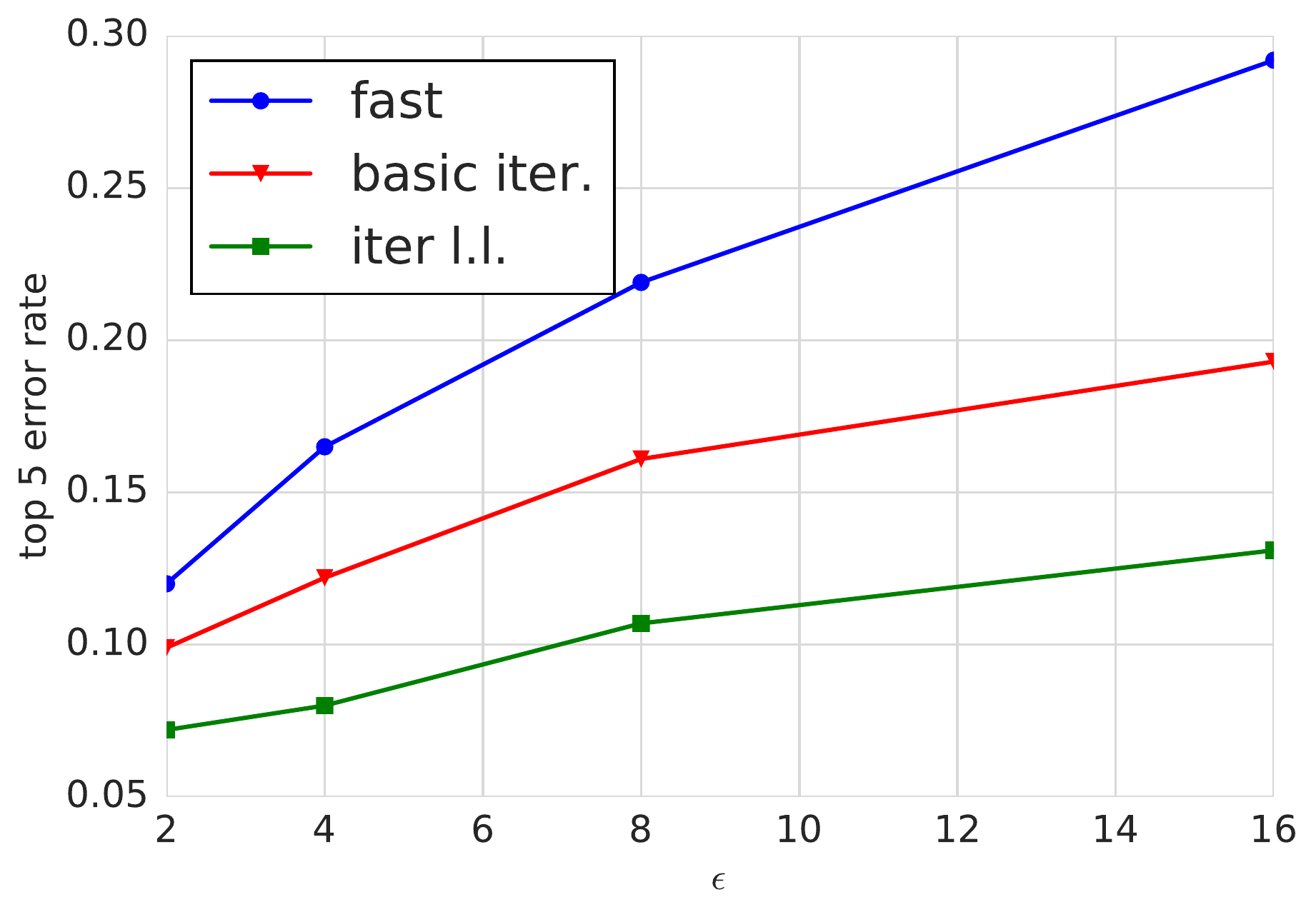}
    \caption{Top 5 error rate.}
  \end{subfigure}
  \caption{Influence of the size of adversarial perturbation on the error rate on adversarial examples
  generated for one model and classified using another model.
  Both source and target models were Inception v3 networks with different random intializations.}\label{fig:transfer-error-rate-eps}
\end{figure}

To compute the error rate we randomly generated $1000$ adversarial images using the source model
and then classified them using the target model.
Results for various models, adversarial methods and fixed $\epsilon=16$ are provided in Table~\ref{table:transfer-error-rate}.
Results for fixed source and target models and various $\epsilon$ are provided in Fig.~\ref{fig:transfer-error-rate-eps}.

Overall the error rate of transferred adversarial examples exhibits
the same behavior as the transfer rate described in Section~\ref{sec:transfer}.

\begin{table}[h]
\caption{Error rates on adversarial examples transferred between models, rounded to the nearest percent.
Results are provided for adversarial images generated using different adversarial methods and fixed perturbation size $\epsilon=16$.
The following models were used for comparison: \textit{A} and \textit{B} are Inception v3 models with different random initializations,
\textit{C} is Inception v3 model with ELU activations instead of Relu, \textit{D} is Inception v4 model.
See also Table \ref{table:transfer-rate} for the transfer rate of adversarial examples, rather than the absolute
error rate.
}
\label{table:transfer-error-rate}
\begin{center}
\begin{tabular}{|l|l|cccc|cccc|cccc|}
\hline
&  & \multicolumn{4}{|c|}{FGSM} & \multicolumn{4}{|c|}{basic iter.} & \multicolumn{4}{|c|}{iter l.l.} \\
\cline{3-14}
& source & \multicolumn{4}{|c|}{target model} & \multicolumn{4}{|c|}{target model} & \multicolumn{4}{|c|}{target model} \\
& model & A & B & C & D & A & B & C & D & A & B & C & D \\
\hline
top 1 & A (v3)     &     65 &    52 &  53 & 45 &          78 &    51 &  50 & 42 &           100 &    32 &  31 & 27 \\
      & B (v3)     &     52 &    66 &  54 & 48 &          50 &    79 &  51 & 43 &            35 &    99 &  34 & 29 \\
      & C (v3 ELU) &     53 &    55 &  70 & 50 &          47 &    46 &  74 & 40 &            31 &    30 & 100 & 28 \\
      & D (v4)     &     47 &    51 &  49 & 62 &          43 &    46 &  45 & 73 &            30 &    31 &  31 & 99 \\
\hline
top 5 & A (v3)     &     46 &    28 &  28 & 22 &          76 &    17 &  18 & 13 &            94 &    12 &  12 &  9 \\
      & B (v3)     &     29 &    46 &  30 & 22 &          19 &    76 &  18 & 16 &            13 &    96 &  12 & 11 \\
      & C (v3 ELU) &     28 &    29 &  55 & 25 &          18 &    19 &  74 & 15 &            12 &    12 &  96 &  9 \\
      & D (v4)     &     23 &    22 &  25 & 40 &          14 &    16 &  16 & 70 &            11 &    11 &  11 & 97 \\
\hline
\end{tabular}
\end{center}
\end{table}

\section{Results with different activation functions}\label{app:activations}

We evaluated robustness to adversarial examples when the network was trained using various non-linear activation functions
instead of the standard $relu$ activation when used with adversarial training on ``step l.l.'' adversarial images.
We tried to use following activation functions instead of $relu$:
\begin{itemize}
\item $tanh(x)$
\item $relu6(x) = min(relu(x), 6)$
\item $ReluDecay_{\beta}(x) = \frac{relu(x)}{1 + \beta relu(x)^2}$ for $\beta \in \{0.1, 0.01, 0.001\}$
\end{itemize}

Training converged using all of these activations, however test performance was not necessarily the same as with $relu$.

$tanh$ and $ReluDecay_{\beta=0.1}$ lose about $2\%$-$3\%$ of accuracy on clean examples and about $10\%$-$20\%$ on ``step l.l.'' adversarial examples.
$relu6$, $ReluDecay_{\beta=0.01}$ and $ReluDecay_{\beta=0.001}$ demonstrated similar accuracy (within $\pm 1\%$) to $relu$ on clean images
and few percent loss of accuracy on ``step l.l.'' images.
At the same time all non-linear activation functions increased classification accuracy on some of the iterative adversarial images.
Detailed results are provided in Table~\ref{table:activation-functions}.

Overall non linear activation functions could be used as an additional measure of defense against iterative adversarial images.

\begin{table}[h]
\caption{Activation functions and robustness to adversarial examples.
For each activation function we adversarially trained the network on ``step l.l.'' adversarial images
and then run classification of clean images and adversarial images generated using various adversarial methods
and $\epsilon$.}
\label{table:activation-functions}
\begin{center}
\begin{tabular}{|l|l|r|r|r|r|r|}
\hline
Adv. method & Activation          &  Clean &  $\epsilon=2$ & $\epsilon=4$ & $\epsilon=8$ & $\epsilon=16$ \\
\hline
Step l.l.   & $relu$              &  77.5\% &       74.6\% &       75.1\% &       75.5\% &       74.5\% \\
            & $relu6$             &  77.7\% &       71.8\% &       73.5\% &       74.5\% &       74.0\% \\
            & $ReluDecay_{0.001}$ &  78.0\% &       74.0\% &       74.9\% &       75.2\% &       73.9\% \\
            & $ReluDecay_{0.01}$  &  77.4\% &       73.6\% &       74.6\% &       75.0\% &       73.6\% \\
            & $ReluDecay_{0.1}$   &  75.3\% &       67.5\% &       67.5\% &       67.0\% &       64.8\% \\
            & $tanh$              &  74.5\% &       63.7\% &       65.1\% &       65.8\% &       61.9\% \\
\hline
Iter. l.l.  & $relu$              &  77.5\% &       30.2\% &        8.0\% &        3.1\% &        1.6\% \\
            & $relu6$             &  77.7\% &       39.8\% &       13.7\% &        4.1\% &        1.9\% \\
            & $ReluDecay_{0.001}$ &  78.0\% &       39.9\% &       12.6\% &        3.8\% &        1.8\% \\
            & $ReluDecay_{0.01}$  &  77.4\% &       36.2\% &       11.2\% &        3.2\% &        1.6\% \\
            & $ReluDecay_{0.1}$   &  75.3\% &       47.0\% &       25.8\% &        6.5\% &        2.4\% \\
            & $tanh$              &  74.5\% &       35.8\% &        6.6\% &        2.7\% &        0.9\% \\
\hline
Basic iter. & $relu$              &  77.5\% &       28.4\% &       23.2\% &       21.5\% &       21.0\% \\
            & $relu6$             &  77.7\% &       31.2\% &       26.1\% &       23.8\% &       23.2\% \\
            & $ReluDecay_{0.001}$ &  78.0\% &       32.9\% &       27.2\% &       24.7\% &       24.1\% \\
            & $ReluDecay_{0.01}$  &  77.4\% &       30.0\% &       24.2\% &       21.4\% &       20.5\% \\
            & $ReluDecay_{0.1}$   &  75.3\% &       26.7\% &       20.6\% &       16.5\% &       15.2\% \\
            & $tanh$              &  74.5\% &       24.5\% &       22.0\% &       20.9\% &       20.7\% \\
\hline
\end{tabular}
\end{center}
\end{table}

\section{Results with different number of adversarial examples in the minibatch}\label{app:num-adv-minibatch}

We studied how number of adversarial examples $k$ in the minibatch affect accuracy on clean and adversarial examples.
Results are summarized in Table~\ref{table:num-adv-in-minibatch}.

Overall we noticed that increase of $k$ lead to increase of accuracy on adversarial examples
and to decrease of accuracy on clean examples.
At the same having more than half of adversarial examples in the minibatch (which correspond to $k>16$ in our case)
does not provide significant improvement of accuracy on adversarial images,
however lead to up to $1\%$ of additional decrease of accuracy on clean images.
Thus for most experiments in the paper we have chosen $k=16$ as a reasonable trade-off between accuracy on clean and
adversarial images.

\begin{table}[h]
\caption{
Results of adversarial training depending on $k$~--- number of adversarial examples in the minibatch.
Adversarial examples for training and evaluation were generated using step l.l. method.\\
Row `No adv` is a baseline result without adversarial training (which is equivalent to $k=0$).\\
Rows `Adv, $k=X$` are results of adversarial training with $X$ adversarial examples in the minibatch.
Total minibatch size is $32$, thus $k=32$ correspond to minibatch without clean examples.
}
\label{table:num-adv-in-minibatch}
\begin{center}
\begin{tabular}{|l|r|r|r|r|r|}
\hline
            &  Clean  & $\epsilon=2$ & $\epsilon=4$ & $\epsilon=8$ & $\epsilon=16$ \\
\hline
No adv      &  78.2\% &   31.5\% &   27.7\% &   27.8\% &   29.7\% \\
Adv, $k=4$  &  78.3\% &   71.7\% &   71.3\% &   69.4\% &   65.8\% \\
Adv, $k=8$  &  78.1\% &   73.2\% &   73.2\% &   72.6\% &   70.5\% \\
Adv, $k=16$ &  77.6\% &   73.8\% &   75.3\% &   76.1\% &   75.4\% \\
Adv, $k=24$ &  77.1\% &   73.0\% &   75.3\% &   76.2\% &   76.0\% \\
Adv, $k=32$ &  76.3\% &   73.4\% &   75.1\% &   75.9\% &   75.8\% \\
\hline
\end{tabular}
\end{center}
\end{table}

\end{appendices}

\end{document}